\documentclass{article}

\usepackage{arxiv}

\usepackage[utf8]{inputenc} 
\usepackage[T1]{fontenc}    
\usepackage{hyperref}       
\usepackage{url}            
\usepackage{booktabs}       
\usepackage{amsfonts}       
\usepackage{nicefrac}       
\usepackage{microtype}      
\usepackage{lipsum}		
\usepackage{graphicx}
\usepackage{natbib}
\usepackage{doi}

\usepackage{courier}  
\usepackage{natbib}  
\usepackage{caption} 
\usepackage[ruled, vlined]{algorithm2e}
\usepackage{array}
\usepackage{amsmath}
\usepackage{bm}
\usepackage{booktabs}

\usepackage{enumitem}
\usepackage{graphicx}
\usepackage{multirow}
\usepackage{natbib}
\usepackage{soul}
\usepackage{xcolor}
\usepackage{caption}
\usepackage{subcaption}
\usepackage{mathtools}

\newcommand{\prob}{\mathbb{P}}
\newcommand{\lam}{\lambda}

\newcommand{\vlam}{\boldsymbol{\lam}}

\newcommand{\vthe}{\boldsymbol{\theta}}

\renewcommand{\prob}{{\mathbb{P}}}
\newcommand{\mT}{\mathcal{T}}

\DeclareMathOperator*{\argmin}{arg\!min}

\title{
Multi-Objective Model-based Reinforcement Learning for Infectious Disease Control
}

\date{} 					

\author{%
  Runzhe Wan \;\; Xinyu Zhang \;\; Rui Song\\
  Department of Statistics\\
  North Carolina State University\\
  \texttt{\{rwan, xzhang97, rsong\}@ncsu.edu} \\
}

\renewcommand{\headeright}{}
\renewcommand{\undertitle}{}

\renewcommand{\shorttitle}{
Multi-Objective Model-based Reinforcement Learning for Infectious Disease Control
}

\hypersetup{
pdftitle={A template for the arxiv style},
pdfsubject={q-bio.NC, q-bio.QM},
pdfauthor={David S.~Hippocampus, Elias D.~Striatum},
pdfkeywords={First keyword, Second keyword, More},
}

\usepackage[ruled,vlined]{algorithm2e}

\usepackage{natbib}
\setcitestyle{authoryear,round,citesep={;},aysep={,},yysep={;}}

\definecolor{mydarkblue}{rgb}{0,0.08,0.45}
\hypersetup{ %
colorlinks=true,
linkcolor=mydarkblue,
citecolor=mydarkblue,
filecolor=mydarkblue,
urlcolor=mydarkblue,
}
\renewcommand{\cite}[1]{\citep{#1}}

\begin{document}
\maketitle

\let\svthefootnote\thefootnote
\newcommand\freefootnote[1]{%
  \let\thefootnote\relax%
  \footnotetext{#1}%
  \let\thefootnote\svthefootnote%
}

\freefootnote{This paper is accepted at the 27th ACM SIGKDD Conference on Knowledge Discovery and Data Mining (KDD 2021). The authors are grateful to the anonymous reviewers for valuable comments and suggestions.}

\begin{abstract}
Severe infectious diseases such as the novel coronavirus (COVID-19) pose a huge threat to public health. 
Stringent control measures, such as school closures and stay-at-home orders, while having significant effects, also bring huge economic losses. 
In the face of an emerging infectious disease, 
a crucial question for policymakers is how to make the trade-off and implement the appropriate interventions timely given the huge uncertainty. 
In this work, we propose a Multi-Objective Model-based Reinforcement Learning framework to facilitate data-driven decision-making and minimize the overall long-term cost. 
Specifically, at each decision point, a Bayesian epidemiological model is first learned as the environment model, and then the proposed model-based multi-objective planning algorithm is applied to find a set of Pareto-optimal policies. 
This framework, combined with the prediction bands for each policy,  provides a real-time decision support tool for policymakers. 
The application is demonstrated with the spread of COVID-19 in China. 
\end{abstract}


\maketitle

\section{Introduction}\label{sec:intro}
The novel coronavirus (COVID-19) has spread rapidly and 
posed a tremendous threat to the global public health \cite{world2020coronavirus}. 
Among the efforts to contain its spread, 
several strict control measures, including school closures and workplace shutdowns, 
have shown high effectiveness \cite{anderson2020will}. 
Nevertheless, these measures bring enormous costs to economies and other public welfare aspects at the same time \cite{eichenbaum2020macroeconomics}. 
For example, an unprecedented unemployment rate in the United States partially caused by some COVID-19 control measures is anticipated by economists \cite{gangopadhyaya2020unemployment}. 
In the face of an emerging infectious disease, 
there are usually multiple objectives conflicting with each other, and either overreaction or under-reaction may result in a substantial unnecessary loss. 
A crucial question for policymakers all around the world is 
\textit{how to make the trade-off and intervene at the right time and in the right amount to minimize the overall long-term cost to the citizens.}

This work is motivated by the ongoing COVID-19 pandemic, where it is witnessed that the decision-making can be impeded by huge intervention costs, great uncertainty on the infectious ability of the disease and the effectiveness of different interventions, and various concerns over the long-term impacts. 
This paper aims to provide a real-time data-driven decision support framework for policymakers. 
We formalize the problem under the multi-objective Markov decision process (MOMDP) framework and integrate epidemiology models, statistical methods, and reinforcement learning algorithms to address this problem.

\textbf{Contribution.  }
Our contributions are multi-fold. 
First, as our transition model, 
we generalize the celebrated Susceptible-Infected-Removal (SIR) model \cite{kermack1927contribution} to allow simultaneous estimation of the infectious ability of this disease and evaluation for the effectiveness of different control measures, in an online fashion. 
There is a vast literature on modeling and prediction for infectious diseases; see \cite{keeling2011modeling} for an overview. 
Among these works, compartmental models such as the SIR model are widely used; see, e.g., \cite{song2020epidemiological} and \cite{sun2020tracking} for applications to COVID-19.  
During an outbreak, the knowledge about the infectious ability and the effectiveness of interventions usually change a lot,
and a quantitative decision-making support tool should utilize the update-to-date information timely. 
However, none of these works considers online parameter updating and intervention effect estimation at the same time. 
We aim to achieve this goal via an online Bayesian framework. 

Second, we propose a novel online planning framework to assist policymakers in making decisions that minimize the overall long-term cost. 
In real applications, policymakers generally group different interventions into several ordered levels with increasing strictness and choose among them subsequently (e.g., in the U.S. \cite{korevaar2020quantifying} and New Zealand \cite{wilson2020pandemic}). 
This paper will focus on selecting among such a set of ordinal actions. 
Specifically, at each decision point, on the ground of the estimated generalized SIR model, 
we propose a model-based planning algorithm to learn the optimal intervention policy, for each given weight between the two competing objectives, i.e., minimizing the epidemiological cost and the economic cost. 
Both interpretable policy classes and complex black-box classes are considered. 
We then extend the algorithm to obtain a representative set of Pareto-optimal policies, i.e., policies that cannot be improved for one objective without sacrificing another.
This framework, combined with the prediction bands for each policy, achieves the goal of supporting multi-objective decision-making. 
Compared with the huge literature on infectious disease modeling and prediction, 
the optimal decision-making problem is far less studied. 
Most works either focus on the evaluation of several fixed interventions 
\cite{tildesley2006optimal, ferguson2020impact, hellewell2020feasibility} 
or study the optimal control problem with a deterministic model \cite{ledzewicz2011optimal, elhia2013optimal}. 
None of these works allows sequential decision making with online updated parameter estimation,
which enables selecting the appropriate intervention according to the current state and available data. 
In addition, the long-term effect is particularly important in this application due to the spread nature of the pandemic. 
Reinforcement learning (RL) is particularly suitable for these purposes, while its application in infectious disease control is relatively new to the literature.
The existing works are mainly concerned with problems caused by limited resources \cite{probert2018real, laber2018optimal}, such as vaccine limitations in controlling seasonal flu. 
In contrast, we focus on another aspect, the multi-objective problem caused by the huge costs of stringent control measures. 
This approach provides us with a clearer view of the trade-off. 
To our knowledge, this is the first work on applications of multi-objective RL to infectious disease control. 

Third, we present an application of our method to control the outbreak of COVID-19 as an example, which is important in its own right. 
Our proposed framework is generally applicable to pandemic controls. 
To investigate the robustness, applications to other diseases and several sensitivity analyses are also studied.

\textbf{Related work.  } 
In addition to the aforementioned literature on infectious disease-related modeling, our methodology also belongs to the field of reinforcement learning. 
Our problem is closely related to the line of research on MOMDPs (see \cite{roijers2013survey} or \cite{liu2014multiobjective} for a survey), which studies problems with multiple competing objectives. 
When the weights between objectives are unknown, the literature focuses on obtaining the whole class of Pareto-optimal policies approximately \cite{castelletti2012tree, barrett2008learning, parisi2017manifold, pirotta2015multi}. 
Because of the online planning nature in our case, it suffices to adopt a simpler approach, by performing multiple runs of policy search over a representative set of weights \cite{van2013scalarized, natarajan2005dynamic}. 

Besides, RL algorithms are commonly classified as model-based methods (e.g., MBVE \cite{feinberg2018model}, MCTS \cite{browne2012survey}, etc.), which directly learn a model of the system dynamics, and model-free methods (e.g., fitted-Q iteration \cite{riedmiller2005neural}, deep-Q network \cite{mnih2015human}, 
actor-critic \cite{konda2000actor}, etc.), which do not. 
In the case of emerging infectious disease control, on one hand, the algorithm needs to generalize to unseen transitions (e.g., the end of the epidemic), where the model-free approach is typically not applicable \cite{van2019use}; 
on the other hand, some epidemiological models have demonstrated satisfactory prediction power. 
Therefore, the model-based approach is adopted in this paper. 

Finally, we note that some efforts have been made in the literature to study the control policy for COVID-19 \cite{alvarez2020simple,piguillem2020optimal,eftekhari2020markovian}. 
As discussed above, these works did not consider the multi-objective problem. 
In addition, they focus on solving a one-time planning problem to decide all future actions instead of providing a real-time decision-making framework. 
RL algorithms have also been applied to learn the 
vaccine distribution strategy \cite{awasthi2020vacsim} or to control mobility \cite{song2020reinforced}. 







\setlength{\textfloatsep}{1cm plus 0.5cm minus 0.5cm}

\textbf{Outline.  }
The remainder of this paper is structured as follows. 
We first introduce some background of 
sequential decision-making and epidemiological modeling in Section \ref{sec:Preliminary}.     
The proposed decision-making workflow is outlined in Section \ref{sec:workflow}, with details of several components discussed in Section  \ref{sec:details}. 
The numerical experiments are presented in Section \ref{sec:numerical}. 
We conclude the paper with discussions and possible extensions in Section \ref{sec:discussion}. 

\section{Preliminary}\label{sec:Preliminary}
\subsection{Multi-objective Markov decision process}\label{sec:pre_MOMDP}
A Markov decision process (MDP) \cite{puterman2014markov} is a sequential decision making model which can be represented by a tuple $\langle \mathcal{S}, \mathcal{A}, c, f \rangle$, where 
$\mathcal{S}$ is the state space, 
$\mathcal{A}$ is the action space, 
$c : \mathcal{S} \times \mathcal{A} \rightarrow \mathbb{R}$ is the expected cost function, 
and $f : \mathcal{S}^2 \times \mathcal{A}  \rightarrow \mathbb{R}$ is a Markov transition kernel. 
Throughout the paper we use the term \textit{cost} instead of \textit{reward}. 
A multi-objective MDP (MOMDP) is an extension of the MDP model when there are several competing objectives. 
An MOMDP can be represented as a tuple $\langle \mathcal{S}, \mathcal{A}, \mathbf{c}, f \rangle$, where the other components are defined as above and $\mathbf{c} = (c^1, \dots, c^K)^T$ is a vector of $K$ cost functions for $K$ different objectives, respectively. 

In this paper, we consider the finite horizon setting with a pre-specified horizon $T$. 
The horizon can simply be selected as a large enough number without loss of generality, since the costs we consider will be close to zero after the disease being controlled. 
Besides, we focus on deterministic policies for realistic consideration, 
since to assign an intervention by randomization may not be feasible. 
For a deterministic policy $\pi$, we define its $k$-th value function at time $t_0$ as $V_{k, t_0}^\pi(s) =  \mathbb{E}_{\pi}( \sum_{t=t_0}^{T} 
c^k(S_t, A_t)|S_{t_0} = s)$, where $\mathbb{E}_{\pi}$ denotes the expectation assuming $A_{t} = \pi(S_t)$ for every $t \ge t_0$. 

In the MDP setting, the objective is typically to find an optimal policy $\pi^*$ that minimizes the expected cumulative cost among a policy class $\mathcal{F}$.
However, in an MOMDP, there may not be a single policy that minimizes all costs. 
Instead, we consider the set of Pareto-optimal policies with linear preference functions $\Pi_{t_0} = \{\pi \in \mathcal{F} : \exists \boldsymbol{\omega} \in \boldsymbol{\Omega} \; \text{s.t.} \; 
\boldsymbol{\omega}^T \boldsymbol{V}_{t_0}^{\pi}(s) \le \boldsymbol{\omega}^T \boldsymbol{V}_{t_0}^{\pi'}(s), 
\forall \pi' \in \mathcal{F}, s \in \mathcal{S} \}$, 
where $\boldsymbol{\Omega} = \{\boldsymbol{\omega} \in \mathbb{R}_+^{K}: \boldsymbol{\omega}^T \boldsymbol{1} = 1\}$ and $\boldsymbol{V}_{t_0}^{\pi} = (V_{1, t_0}^\pi, \dots, V_{K, t_0}^\pi)^T$. 
A policy is called \textit{Pareto-optimal} if it can not be improved for one objective without sacrificing the others. 
The dependency on $t_0$ is to be consistent with the online planning setting considered in this paper.

\subsection{The Susceptible-Infected-Removal model}\label{sec:SIR}
The compartmental models are, arguably, the most popular choices in epidemiology modeling \cite{tang2020review}. 
These models will divide the population into different compartments with labels, and model the transitions of people between these compartments to describe the mechanism of infectious diseases spread. 

\begin{figure}[!h]
\centering
\includegraphics[width=.5\textwidth]{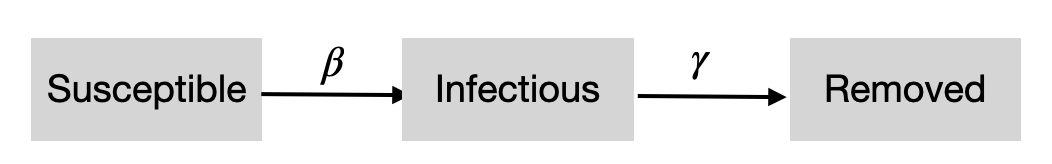}
\caption{The infectious disease spread mechanism described by the SIR model.}
\label{fig:SIR_diagram}
\end{figure}

Among these models, the Susceptible-Infected-Removal (SIR) model is one of the most widely applied models \cite{brauer2008compartmental}. 
Suppose at time $t$, the infectious disease is spread within $N_t$ regions. Denote the total population of the $l$-th region as $M_l$, for $l=1,\dots, N_t$. 
With the SIR model, we divide the total population into three groups: the individuals who can infect others, who have been removed from the infection system, and  who have not been infected and are still susceptible. 
Denote the count for each group in region $l$ at time $t$ as $X^I_{l,t}$, $X^R_{l,t}$, and $X^S_{l,t} = M_l - X^I_{l,t} - X^R_{l,t}$, respectively. 
We will discuss how to construct these variables using surveillance data   in Section \ref{sec:delayed}. 
The standard deterministic SIR model can then be written as 
a system of difference equations: 
\begin{align*}
    X^S_{l,t+1} &= X^S_{l,t} - \beta X^S_{l,t} X^I_{l,t}/ M_l;\\
    X^I_{l,t+1} &= X^I_{l,t} + \beta X^S_{l,t} X^I_{l,t}/ M_l
    -  \gamma X^I_{l,t};\\
    X^R_{l,t+1} &= X^R_{l,t} +  \gamma X^I_{l,t}, 
\end{align*}
where $\beta$ and $\gamma$ are constants representing the infection and removal rate, respectively. 
A diagram for the SIR model is given in Figure \ref{fig:SIR_diagram}.

\section{Framework of the Decision Support Tool}\label{sec:workflow}
In this section, we introduce the main framework. 
The transition model is first introduced in Section \ref{sec:GSIR}, and then the sequential decision-making problem for a given weight is  defined in Section \ref{sec:MDP}, with the multi-objective problem addressed in Section \ref{sec:Pareto}.

\subsection{The generalized SIR model}\label{sec:GSIR}
In this work, we aim to modify the SIR model as our transition model. 
We note that some other compartmental models such as the SEIR model  \cite{brauer2008compartmental} can also fit in our framework. 
However, with limited data and knowledge in the face of an emerging infectious disease, the SIR model has been found to be a robust choice \cite{roda2020difficult}. 

There are two limitations with the standard SIR model. 
First, the infection rate heavily depends on the control measure being taken. 
As discussed in Section \ref{sec:intro}, in this paper we focus on an ordinal set of actions  $\mathcal{A} = \{1, 2, \dots, J\}$, with level $1$ standing for no official measures. 
Second, a stochastic model with proper distribution assumptions is required to fit real data with randomness. 
Motivated by the discussions above, we propose the generalized SIR (GSIR) model and use it as our \textbf{transition model}:

\begin{align}
\label{SIR}
\begin{split}
X^S_{l,t+1} &= X^S_{l,t} - e^S_{l,t}, \;\;  e^S_{l,t} \sim  \text{Poisson}(\sum_{j=1}^J \beta_j \mathbb{I}(A_{l,t} = j) X^S_{l,t} \frac{X^I_{l,t}}{M_l} ) ;\\
X^R_{l,t+1 } &= X^R_{l,t} + e^R_{l,t}, \;\;  e^R_{l,t} \sim \text{Binomial}(X^I_{l,t}, \gamma );\\
X^I_{l,t+1} &= M_l - X^S_{l,t+1} - X^R_{l,t+1 }, 
\end{split}
\end{align}
where $\mathbb{I}(\cdot)$ is the indicator function, 
$A_{l,t}$ is the action taken by region $l$ at time $t$, 
and  $\beta_j$ denotes the infection rate under action $j$. 
The choice of the Poisson and the Binomial distribution is popular in the literature \citep{held2019handbook}, and is motivated by their probabilistic
implications. 
We assume $\beta_1 \ge \beta_2 \ge \dots \ge \beta_J \ge 0$ to represent the increasing effectiveness. 
The estimation of $\vthe = (\gamma, \beta_1, \dots, \beta_J)^T$ is deferred to Section \ref{sec:sir_est}.





\begin{figure*}[!t]
\centering
\includegraphics[width=\textwidth, height=.3\textwidth]{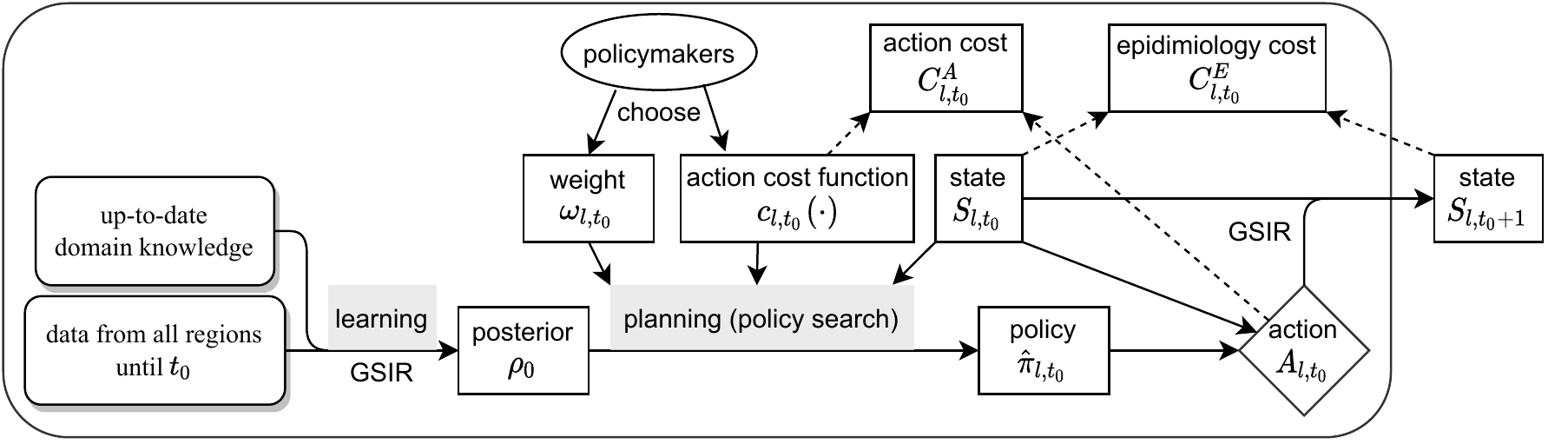}
\caption{Decision-making process for region $l$ at time $t_0$}
\label{fig:diagram}
\end{figure*}

\subsection{Sequential decision making}\label{sec:MDP}
In this work, we focus on the intervention decision of each region: 
in the model estimation (learning) step, data from several similar regions are 
aggregated to share information and mitigate the issue that data is typically noisy, scarce, and single-episode in a pandemic; 
while in the decision-making (planning) step, each region chooses its own action.

\textbf{MOMDP definition.  }
The intervention decision-making problem can be naturally formalized as an MOMDP.
For each region $l$, 
at each decision point $t$, according to the estimated transition model, the current  \textbf{state} $S_{l,t} = (X^S_{l,t}, X^I_{l,t}, X^R_{l,t})^T$, and their judgment, 
the policymakers determine a \textbf{policy} $\pi_{l,t}$ with the objective of  minimizing the overall long-term cost, 
and  choose the \textbf{action} $A_{l,t}$ to implement according to the policy. 
This action, assuming being effectively executed, will affect the infection rate during $(t, t+1]$ and hence the conditional distribution of $S_{l,t+1}$. 
Let $f(\cdot|\cdot, \cdot; \vthe)$ be the 
conditional density for $S_{l, t+1}$ given $S_{l,t}$ and $A_{l,t}$ in model (\ref{SIR}). 
In this work, we consider two cost variables, the \textbf{epidemiological cost} $C^E_{l,t}$ and the \textbf{action cost} $C^A_{l,t}$. 
$C^E_{l,t}$ can be naturally chosen as the number of new infections $X^S_{l,t} - X^S_{l,t+1}$. 
Let $C^A_{l,t} = c_{l}(Z_{l,t}, A_{l,t}, Z_{l,t+1}) \in \mathbb{R}^+$ for a time-varying  variable  $Z_{l,t}$ and a stochastic function $c_l$. 
Both $c_l$ and  $Z_{l,t}$ should be chosen by domain experts. 
For example, with $Z_{l,t}$ representing the unemployment rate, $C^A_{l,t} = Z_{l,t + 1} - Z_{l,t}$ can represent its change due to $A_{l,t}$. 
Since the modeling for action cost and its transition is a separate question, for simplicity, 
we focus on the case that $C^A_{l,t} = c_l(A_{l,t})$ for a pre-specified  stochastic function $c_l(\cdot)$ in this paper. 
$c_l(\cdot)$ is set to be region-specific to incorporate local features, such as the economic conditions. 
With a given weight $\omega_{l,t} \in \mathbb{R}^+$, the \textbf{overall cost} $C_{l,t}$ is then defined as $C^E_{l,t} + \omega_{l,t} C^A_{l,t}$.
The expected cost functions can then be derived from these definitions, and for decision-making purposes, the overall cost is equivalent with the weighted cost defined in Section  \ref{sec:pre_MOMDP} when $K = 2$.

\textbf{Online planning.  } 
The RL problem considered in this paper is an \textit{online planning}, or sometimes called \textit{planning at decision time} problem \cite{sutton2018reinforcement}, which  focuses on selecting an action for the current state with collected environment information. 
We consider a sequence of decision points $\mathcal{T} \subset \{1, \dots, T\}$ to reduce the action switch cost in real applications. 
At time $t_0 \notin \mathcal{T}$, the region keeps the same action with time $t_0 - 1$. 
At time $t_0 \in  \mathcal{T}$, 
the policymakers choose an action according to the decision-making workflow displayed in Figure \ref{fig:diagram}. 

We summarize Figure \ref{fig:diagram} as follows. 
For each region $l$, at the decision point $t_0$, 
we first estimate the posterior  of $\vthe$ as  $\rho_{t_0}$, using accumulated data $\mathcal{D}_{t_0} {=} \{S_{l,t}, A_{l,t}\}_{1 \le l \le N_{t_0}, 1 \le t \le t_0 - 1} \cup \{S_{l,t_0}\}_{1 \le l \le N_{t_0}}$ and priors selected with domain knowledge. 
Next, the policymakers choose the trade-off weight  $\omega_{l, t_0}$, 
learn a deterministic policy $\hat{\pi}_{l, t_0}(\cdot; \omega_{l, t_0})$ by planning, and implement $A_{l,t_0} {=} \hat{\pi}_{l, t_0}(S_{l,t_0}; \omega_{l, t_0})$. 
Formally, we solve the following optimization problem to obtain $\hat{\pi}_{l, t_0}(\cdot; \omega_{l, t_0})$:
\vspace{-.3cm}
\begin{align}
\hat{\pi}_{l, t_0}(\cdot; \omega_{l, t_0}) = \argmin_{\pi \in \mathcal{F}} \mathbb{E}_{\pi, \rho_{t_0}}( \sum_{t=t_0}^{T}(  C^E_{l,t} + \omega_{l, t_0}  C^A_{l,t})), 
\label{fomular:objective}
\end{align}
where  $\mathbb{E}_{\pi, \rho_{t_0}}$ denotes the expectation assuming 
$C^A_{l,t} \sim c_l(A_{l,t})$, 
$\prob(S_{l, t+1} = s_{l, t+1} |S_{l,t}\! =  s_{l,t}, A_{l,t}\! = a_{l,t} ) \!= f(s_{l, t+1} |s_{l,t}, a_{l,t}; \vthe)$ with $\vthe  \,{\sim}\, \rho_{t_0}$, and 
$A_{l,t} \!= \pi(S_{l,t})\mathbb{I}(t \in \mathcal{T}) \!+ A_{l,t-1} \mathbb{I}(t \notin \mathcal{T})$, 
for every $t {\ge} t_0$. 

The Bayesian approach is preferred because 
(i) we need to make decisions before accumulating sufficient data,  and typically there is important domain knowledge available, and 
(ii) in this important application, the estimation uncertainty should be taken into consideration in decision-making.  
The specification of $\mathcal{F}$ and the policy search algorithms for solving (\ref{fomular:objective}) will be discussed in Section \ref{sec:policy}. 


\subsection{Multiple objectives and Pareto-optimal policies}\label{sec:Pareto}
In the discussion above, 
we assume the tradeoff weight $\omega_{l, t_0}$ is easily specified at each decision point $t_0$. 
This is feasible when the two objectives share the same unit, for example, when $C^A_{l,t}$ represents the damage to public health due to economic losses. 
In general settings, properly choosing the weight is not easy and sometimes unrealistic.
Therefore, we aim to assemble a decision support tool that provides a comprehensive picture of the future possibilities associated with different weight choices and hence makes the multi-objective decision-making feasible.

\begin{figure*}[!t]
\centering
\begin{subfigure}{.23\textwidth}
  \centering
  \includegraphics[width=.9\linewidth]{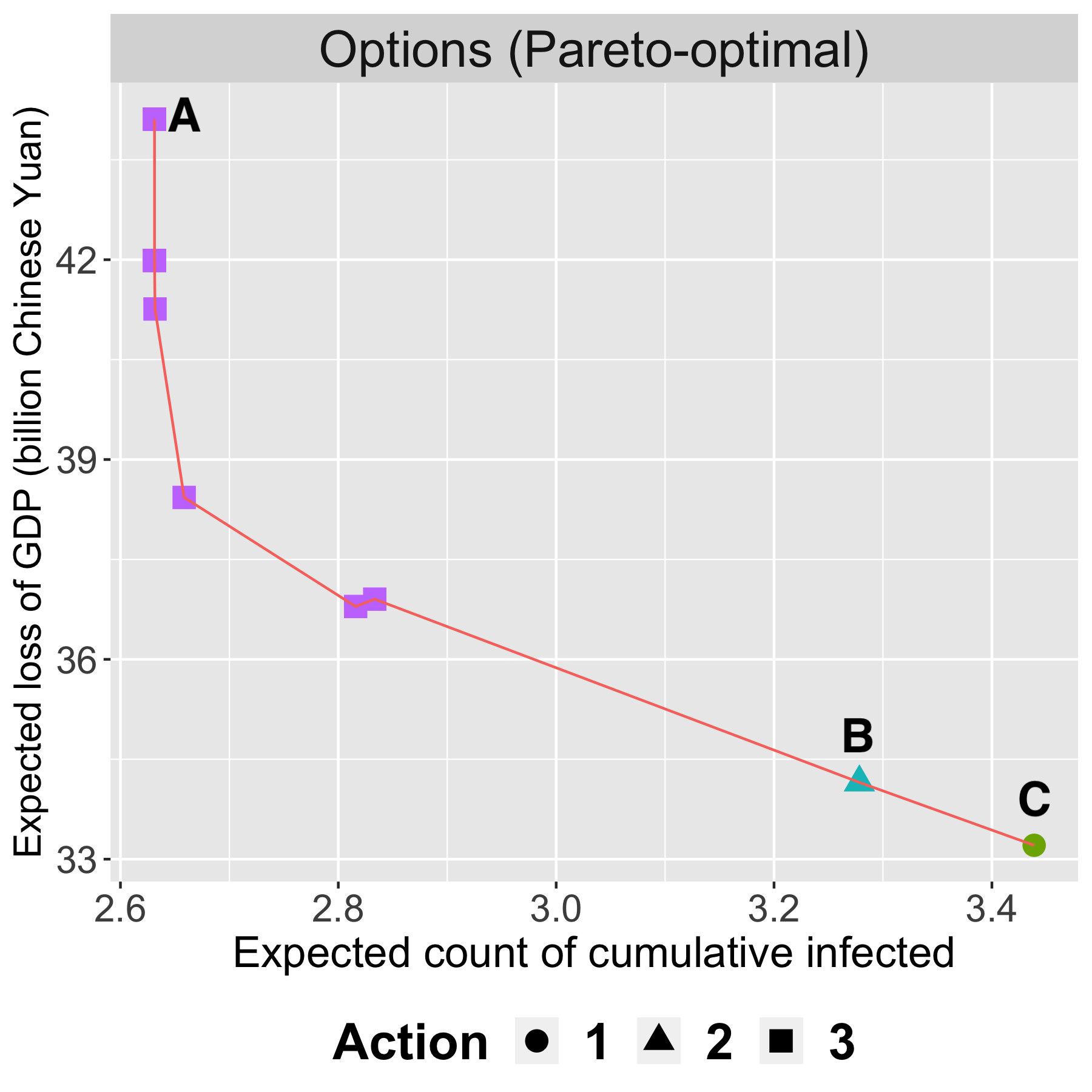}
  \caption{Expected cumulative costs for different policies, with the corresponding recommended immediate actions indicated by the shape. 
  }
\end{subfigure}%
\hspace{.2cm}
\begin{subfigure}{.23\textwidth}
  \centering
  \vspace{-.6cm}
  \includegraphics[width=.9\linewidth]{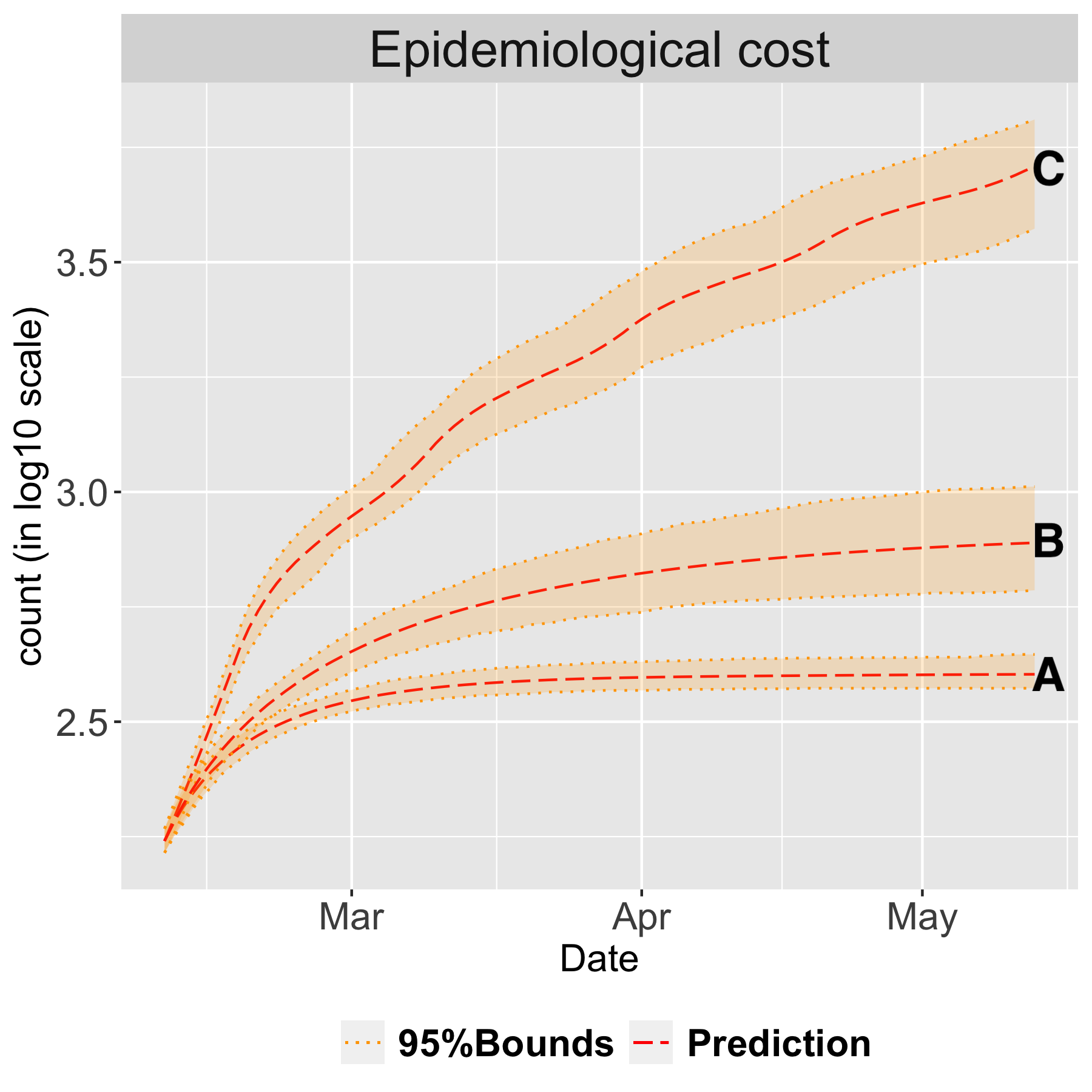}
  \caption{ 
  Mean trajectory and prediction bands of the predicted cumulative epidemiological costs. 
  }
\end{subfigure}%
\hspace{.2cm}
\begin{subfigure}{.23\textwidth}
  \centering
  \vspace{-.6cm}
  \includegraphics[width=.9\linewidth]{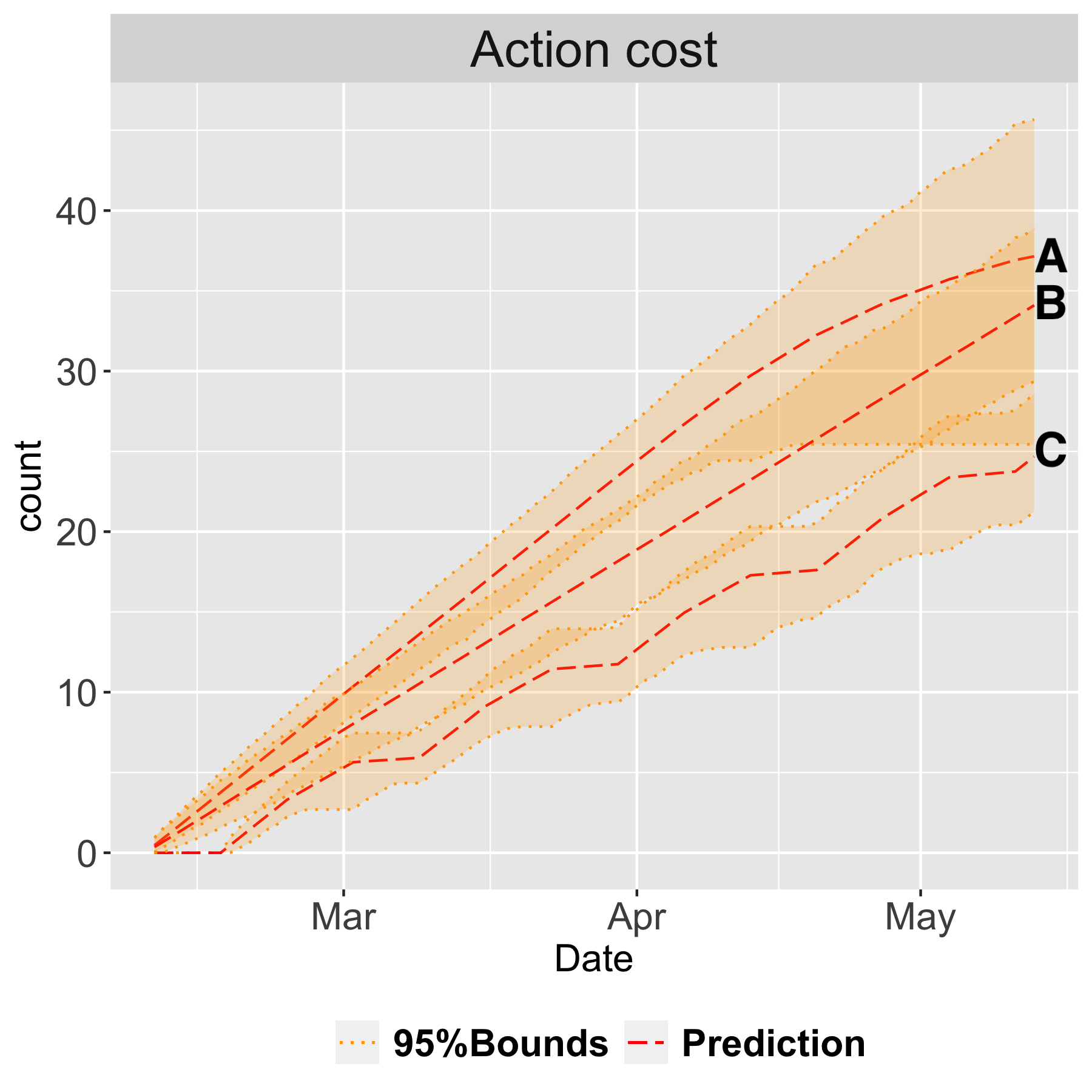}
  \caption{
    Mean trajectory and prediction bands of the predicted cumulative action costs. 
  }
\end{subfigure}%
\hspace{.2cm}
\begin{subfigure}{.23\textwidth}
  \centering
  \includegraphics[width=.9\linewidth]{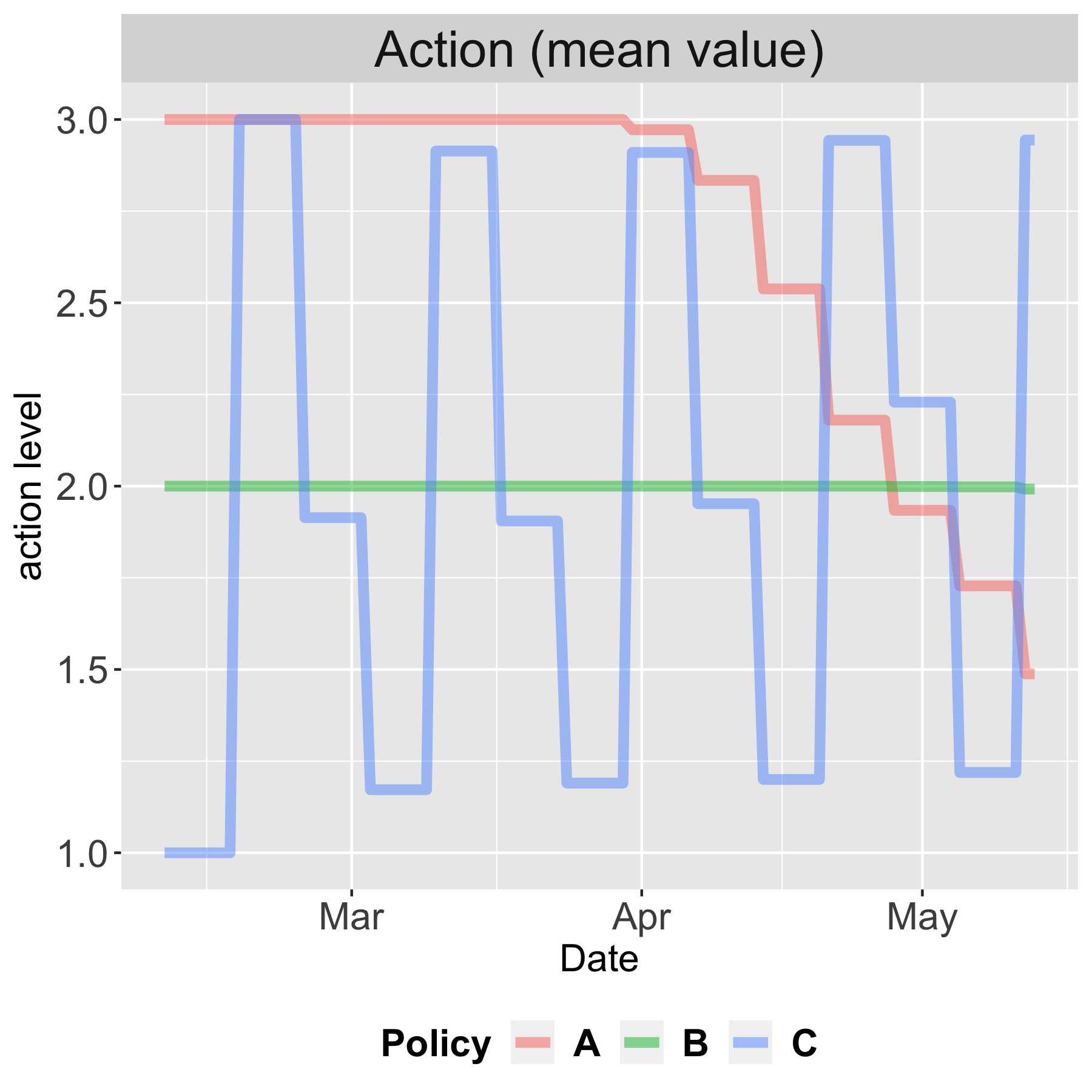}
  \caption{
  Mean trajectory of the predicted recommended actions. 
  Levels are averaged over replications and hence may not be integers. 
  }
\end{subfigure}%

\caption{A demo of some information presented to the users. 
Results are computed for Beijing, China on 02/10/2020. 
The proposed method is run under the experiment setting introduced in Section \ref{sec:numerical}, where we consider three levels of actions. 
For three sample policies, we plot the predicted behaviour of them in subplot (b)-(d). 
Prediction results are aggregated over $1000$ replications. 
According to subplot (d), policies A and B  can be interpreted as the suppression and mitigation policy studied in  \cite{ferguson2020impact}, and  policy C is close to the periodic lockdown policy suggested in the literature \cite{bin2020post, karin2020cyclic}.  
}
\label{fig:pareto_demo}
\end{figure*}

Solving the whole set of Pareto-optimal policies is typically challenging. 
While in our online planning setting, at each decision point $t_0$, 
we only need to make a one-time decision among a few available actions. 
For this purpose, it suffices to solve problem (\ref{fomular:objective}) for a representative set of weights $\{\omega_b\}_{b=1}^B$ to find the corresponding Pareto-optimal policies, and then apply Monte Carlo simulation to obtain the corresponding prediction bands for the potential costs following each policy. 
The policymakers can then compare all these possible options, select among them, and hence choose the action $A_{l, t_0}$. 
With this module, our framework is an \textit{user-in-the-loop} tool. 
A demo of some information presented to the users is displayed in Figure \ref{fig:pareto_demo}. 
We summarize this tool in Algorithm \ref{alg:tool}. 

\begin{algorithm}[ht]
\DontPrintSemicolon
\KwIn{weights $\{\omega_b\}_{b=1}^B$, number of replications $K$, significance level $\alpha$, action cost function $c_l(\cdot)$, decision points $\mathcal{T}$, $\mathcal{D}_{t_0}, \rho_{t_0}, T$.
}

\For{$b = 1, \dots, B$}{
apply a policy search algorithm to find the optimal policy  $\hat{\pi}^b$ for weight $\omega_b$. 

 \For{$k = 1, \dots, K$}{
 set the cumulative cost $V^{k, E}_{t_0 - 1}$ and  $V^{k, A}_{t_0 - 1}$ as $0$
 
 set $S^{k}_{t_0,b} = S_{t_0}$
 
  \For{$t= t_0,\dots,T$}{

            
            choose action $A^k_{t,b} = \hat{\pi}^b(S^{k}_{t,b}) \mathbb{I}(t \in \mathcal{T}) + A^k_{t-1,b} \mathbb{I}(t \notin \mathcal{T})
            $
            
            sample $C^{k, A}_{t,b} \sim c_l(A^k_{t,b})$, 
            $\vthe^k_{t,b} \sim \rho_{t_0}$, and $S^k_{t+1,b} \sim  f(\cdot|S^k_{t,b}, A^k_{t,b}; \vthe^k_{t,b})$
             
            calculate $V^{k, E}_{t,b} = V^{k, E}_{t - 1,b} + X^{k, S}_{t,b} - X^{k, S}_{t + 1,b}$ and $V^{k, A}_{t,b} = V^{k, A}_{t - 1,b} + C^{k, A}_{t,b}$
        }
    }
    for every $t$, calculate the upper and lower 
    $\alpha/2$-th quantile and the mean of $\{V^{k, E}_{t,b}\}_{k=1}^K$ as $V^{b,E}_{u,t}$, $V^{b,E}_{l,t}$, and $\bar{V}^{b,E}_{t}$, respectively; 
    similarly calculate $V^{b,A}_{u,t}$,  $V^{b,A}_{l,t}$, and  $\bar{V}^{b,A}_{t}$.
}

\KwResult{ optimal policies $\{\hat{\pi}^b\}_{b = 1}^B$
, recommended actions $\{A^1_{t_0,b}\}_{b = 1}^B$, 
prediction bands for  costs
$\{\{(V^{b,E}_{l,t}, V^{b,E}_{u,t}, \bar{V}^{b,E}_{t}),
(V^{b,A}_{l,t}, V^{b,A}_{u,t}, \bar{V}^{b,A}_{t}) \}_{t = t_0}^T
\}_{b = 1}^B$ }  
\caption{Pareto-optimal Policies and Prediction Bands
}\label{alg:tool}
\end{algorithm}

\section{Details of the Components}\label{sec:details}
\subsection{State construction}\label{sec:delayed}
In this section, we discuss the construction of state variables with surveillance data. 
The data usually available to policymakers is the cumulative count of confirmed cases until time $t$, denoted as $O^I_{l,t}$. 
Unlike the infectious disease spread in nature, the individuals counted in $O^I_{l,t}$ generally either have been confirmed and isolated, or have recovered or died. 
Therefore, these individuals can be regarded as those removed from the infection system, and it is natural to set $X^R_{l,t}$ as $O^I_{l,t}$ \cite{yang2020sir}.  

Besides, for infectious diseases, there is typically a time delay between being infectious and getting isolated, the length of which is treated as a random variable with expectation $D$. 
Therefore, the count of the infectious $X^I_{l,t}$ is usually not immediately observable at time $t$, but will be gradually identified in the following days. 
Following the existing works on infectious disease modelling \cite{zhang2005compartmental, chen2020data}, 
we treat this issue as a delayed observation problem and use $O^I_{l,t + D} - O^I_{l,t}$, the new confirmed cases during $(t, t+D]$, as a proxy for $X^I_{l,t}$. 
In the planning step, following the literature on delayed MDPs \cite{walsh2009learning}, we apply Algorithm \ref{alg:MBS} (Model-based Simulation, MBS) to generate a proxy state, and then choose the action according to it. 
We note that although this issue is not obvious in prediction, it is unavoidable in  decision-making because $A_{l,t}$ works directly on $X^I_{l,t}$. 
The performance of such an approximation is supported by the theoretical analysis in \cite{walsh2009learning} and 
the experiments in Section \ref{sec:numerical}.

 \begin{algorithm}[ht]
\KwData{$\{ O^I_{l,t}\}_{t = t_0 - D}^{t_0}$, $\{A_{l,t}\}_{t = t_0 - D}^{t_0 - 1}$, $\rho_{t_0}$,  and $D$
}
set $X^{I,G}_{l,t_0 - D} = O^{I}_{l,t_0} - O^{I}_{l,t_0 - D}$ and $(\hat{\gamma}_{t_0}, \hat{\beta}_{1,t_0}, \dots, \hat{\beta}_{J,t_0})^T = \mathbb{E}(\rho_{t_0})$.

\For{$t= t_0-D, \dots, t_0 - 1$}{
  $X^{I,G}_{l,t+1} = X^{I,G}_{l,t} + \sum_{j=1}^J \hat{\beta}_{j, t_0}  \mathbb{I}_{\{A_{l,t} = j\}} (M_l - O^I_{l,t} - X^{I,G}_{l,t}) X^{I,G}_{l,t} / M_l
- (O^I_{l,t + 1} - O^I_{l,t})$
}



\KwResult{the proxy state $S_{l,t_0} = (M_{l} - X^{I, G}_{l,t_0} - O^I_{l,t_0}, X^{I,G}_{l,t_0}, O^I_{l,t_0})^T$}
\caption{Model-based Simulation (MBS)}\label{alg:MBS}
\end{algorithm}

\subsection{Estimation of the transition model}\label{sec:sir_est}
At each time $t_0 \in \mT$, we need to first obtain the posterior of $\vthe$ in the transition model (\ref{SIR}). 
Notice that the last equation in (\ref{SIR}) is  redundant under the constraint $X^S_{l,t} + X^I_{l,t} + X^R_{l,t} = M_l$ for all $t$. 
With data $\mathcal{D}_{t_0}$, 
the Markov property reduces the estimation problem to $J + 1$  Bayesian generalized linear models with the identity link function. 
By choosing the priors as the conjugate priors, the posterior distributions have explicit forms. 
Specifically, let $\gamma \sim Beta(a_{R}, b_{R})$ with parameters $(a_{R}, b_{R})$, and  let $\beta_j \sim Beta(a_{S,j}, b_{S,j})$ with parameters $(a_{S, j}, b_{S, j})$, for $j = 1, \dots, J$. 
Under the assumption that these priors are independent, we have 
$\gamma \mid \mathcal{D}_{t_0} \sim  Beta(a_{R}^* , b_{R}^*)$ and 
$\beta_j \mid \mathcal{D}_{t_0} \sim  Gamma(a_{S,j}^* , b_{S,j}^*)$, for $j = 1, \dots, J$ , where
\begin{align*}
a_{S,j}^* &= a_{S,j} + \smashoperator{\sum_{(t,l) \in \Omega_{t_0,j}}} \left( {X_{l, t}^{S} - X_{l, t+1}^{S}}{} \right),
b_{S,j}^* = b_{S,j}+ \smashoperator{\sum_{(t,l) \in \Omega_{t_0,j}}}  \frac{X_{l, t}^{S}X_{l, t+1}^{I}}{M_l}, \\
a_{R}^* &= a_{R} + \smashoperator{\sum_{(t,l) \in \Omega_{t_0}}} \left( {X_{l, t+1}^{R} - X_{l, t}^{R}} \right), 
b_{R}^* = b_{R}- \smashoperator{\sum_{(t,l) \in \Omega_{t_0}}}  \left( {X_{l, t+1}^{R} - X_{l, t}^{R}} \right) + \smashoperator{\sum_{(t,l) \in \Omega_{t_0}}} {X_{l, t}^{I}}, 
\end{align*}
where  
${\Omega_{t_0,j}} {=}
\{(t,l) : 1 {\leq} t {\leq} t_0 {-} 1,
1 {\leq} l {\leq} N_{t}, 
A_{l,t} {=} j\}$ is the set of time-location pairs where action $j$ is taken, and $\Omega_{t_0} {=} \bigcup_j \Omega_{t_0,j}$. 

Below, we introduce a way to specify the prior parameters: 
(i) $\beta_1$ and $\gamma$ are both features of this disease without any interventions, and we can first set their priors with the estimates of similar diseases, and update them when additional biochemical findings are available;
(ii) for $j \geq 2$, $\beta_j$  indicates the infection rate under action level $j$. 
Suppose we have a reasonable estimate of the intervention effect $u_j= \beta_j / \beta_1$ as $\hat{u}_j$ and that of $\beta_1$ as $ \hat{\beta}_1$, then the prior of $\beta_j$ can be set as a distribution with expectation $\hat{u}_j \hat{\beta}_1$.

\subsection{Policy search}\label{sec:policy}
At each decision point $t_0 \in \mathcal{T}$, for each region $l$ and a given weight $\omega_{l, t_0}$, we need to learn the optimal policy $\hat{\pi}_{l, t_0}(\cdot; \omega_{l, t_0}) \in \mathcal{F}$ by solving the model-based planning problem (\ref{fomular:objective}). 
We introduce two types of policy classes and the corresponding planning  algorithms, with emphasis on either the interpretability or the global optimality.  

\textbf{Interpretable policy class.  } 
In applications of infectious disease control, the interpretability of the policy is typically important. 
In this case, we need to restrict our attention to an interpretable parametric policy class. 
As an illustration, in this work, we consider the following observations from the pandemic control decision-making process in real life: 
(i) the decision should be based on the spread severity, which we interpret as $X^I_{l,t}$, the number of infectious individuals; 
(ii) the policy should also be based on the current estimate of disease features $\rho_{t_0}$, the current state $S_{l, t_0}$, the trade-off weight $\omega_{l,t_0}$, and the potential cost $c_l(\cdot)$, which have all been incorporated in the objective function of (\ref{fomular:objective}); 
(iii) the more serious the situation, the more stringent the intervention should be. 
Motivated by these observations, we focus on the 
threshold-based policy class: 
\begin{equation}\label{eqn: policy_class_thre}
\begin{split}
 \mathcal{F} = \{\pi : \pi(S_{l,t}; \vlam) = \sum_{j=1}^{J} j \mathbb{I}(\lam_j \le X^I_{l,t} < \lam_{j+1}), \\ 0 = \lam_1 \le \lam_2 \le \dots \le \lam_{J + 1} = M_l, \lam_J \le \lam_M
    \}, 
\end{split}
\end{equation}
where $\vlam = (\lam_2,\dots, \lam_{J})^T$.
Notice that $X^I_{l,t}$ is generally a number much smaller than the population $M_l$, 
we introduce the pre-specified tolerance parameter $\lam_{M} \in (0,M_l)$ to reduce the computational cost by deleting unrealistic policies. 

Given  a specified policy class, problem (\ref{fomular:objective}) can be then solved via  rollout-based direct policy search. 
Direct policy search algorithms \cite{gosavi2015simulation} generally apply optimization algorithms to maximize the value function approximated via Monte Carlo rollouts. 
For class (\ref{eqn: policy_class_thre}), since the state transition and the action cost are both computationally affordable to sample, 
when $J$ is not large,  we can simply apply the grid search algorithm, which is  robust and safe in such an important application. 
The example for $J = 3$, which is the case in our experiment, is described in Algorithm \ref{alg:grid} in Appendix \ref{sec:pseudo_code}. 
When $J$ is large, many other optimization algorithms such as the simultaneous perturbation stochastic approximation algorithm  \cite{sadegh1997constrained} can be used.  
Alternatively, one can apply the deterministic policy gradient \citep{silver2014deterministic} or actor-critic 
\citep{lillicrap2015continuous} with the learned environment model as a simulator. 

\textbf{Black-box policy class.  } 
If the interpretability can be traded off for global optimality, a more complex policy class can be considered. 
Since the action set is discrete, multiple planning algorithms in the literature are applicable, such as Monte carlo tree search with state aggregation \cite{hostetler2014state}. 
As an example, in this work, we focus on the policy class $\mathcal{F}$ induced by a class of deep-Q value network (DQN) \cite{mnih2015human} with an augmented state space $\mathcal{S} \times \{1, 2, \dots, T\}$. 
Specifically, to take into consideration the learned model uncertainty $\rho_{t_0}$, we define the Q-function of policy $\pi$ as $Q^\pi_{l, t_0}((s,t'), a; \omega_{l, t_0}) = \mathbb{E}_{\pi, \rho_{t_0}}( \sum_{t=t'}^{T}(  C^E_{l,t} + \omega_{l, t_0}  C^A_{l,t}))$, and then apply value iteration with epsilon-greedy exploration to learn the optimal value function as $\hat{Q}^*_{l, t_0}(\cdot, \cdot; \omega_{l, t_0})$. 
The corresponding policy is the induced greedy policy. 
More details can be found in Appendix \ref{sec:pseudo_code}. 

\section{Experiments}\label{sec:numerical}
In this section, we apply our framework to some COVID-19 data in China for illustration. 
China has passed the first peak, which provides data of good quality for validation purposes. 
To further investigate the applicability of our framework to other diseases and its robustness, we also run an experiment with parameters of the 2009 H1N1 pandemic and conduct several sensitivity analysis.

\vspace{-0.2cm}
\subsection{Data description and hyper-parameters}\label{sec:data}
We collect data for six important cities in China from 01/15/2020 to 05/13/2020, 
and index these cities by $l \in \{1, \dots, 6\}$ and dates by $t \in \{1, \dots, 120\}$, with $T = 120$. 
More details about the dataset, data sources, and hyper-parameters can be found in Appendix \ref{sec:supp-data}. 

\vspace{-0.2cm}
\begin{description}
\item[State variables and region-specific features:] 
For each region $l$, 
we collect its annual gross domestic product (GDP) $G_l$,  population $M_l$, and counts of confirmed cases from official sources. 

\item[Action:] 
three levels of interventions implemented in China during COVID-19 are considered: 
level $1$ means no or few official policies claimed; 
level $2$ means the public health emergency response;  
level $3$ means the stringent closed-off management required by the government. 
Data are collected from the news. 

\item[Cost:]  we use $r_{l,t}$, the observed ratio of human mobility loss in city $l$ on day $t$ compared with year 2019, to construct a proxy for its GDP loss and calibrate the action cost function $c_l(\cdot)$. 
For $j \in \{2, 3\}$, we first fit a normal distribution $\mathcal{N}(\mu_j,\sigma_{j}^2)$ to the observed loss ratios under the corresponding intervention level. 
We then define $c_l(a) $ as $ \sum_{j=1}^3 C_j \mathbb{I}(a = j) G_l / 365$, where $C_1 {=} 0$ and $C_j {\sim} \mathcal{N}(\mu_j,\sigma_{j}^2)$ for $j \in \{2, 3\}$. 
We note that this is only for illustration purposes. In real applications, policymakers  need  to carefully design and measure the potential costs with domain experts. 
\item[Hyper-parameters:] 
the parameter estimates for a similar pandemic SARS  \cite{mkhatshwa2010modeling} are used as priors of $\gamma$ and $\beta_1$. 
Similar to \cite{ferguson2020impact}, we assume that action $2$ and $3$ can reduce the infection rate by $80\%$ and $90\%$, respectively, and set the priors for $\beta_2$ and $\beta_3$ accordingly. 
$D$ is chosen as $9$ according to \cite{sun2020tracking} and \cite{pellis2020challenges}. 
\end{description}

\subsection{Estimation and validation of the transition model}\label{sec:numerical_SIR}
\begin{figure}[!h]
\centering
\includegraphics[width=.8\textwidth]{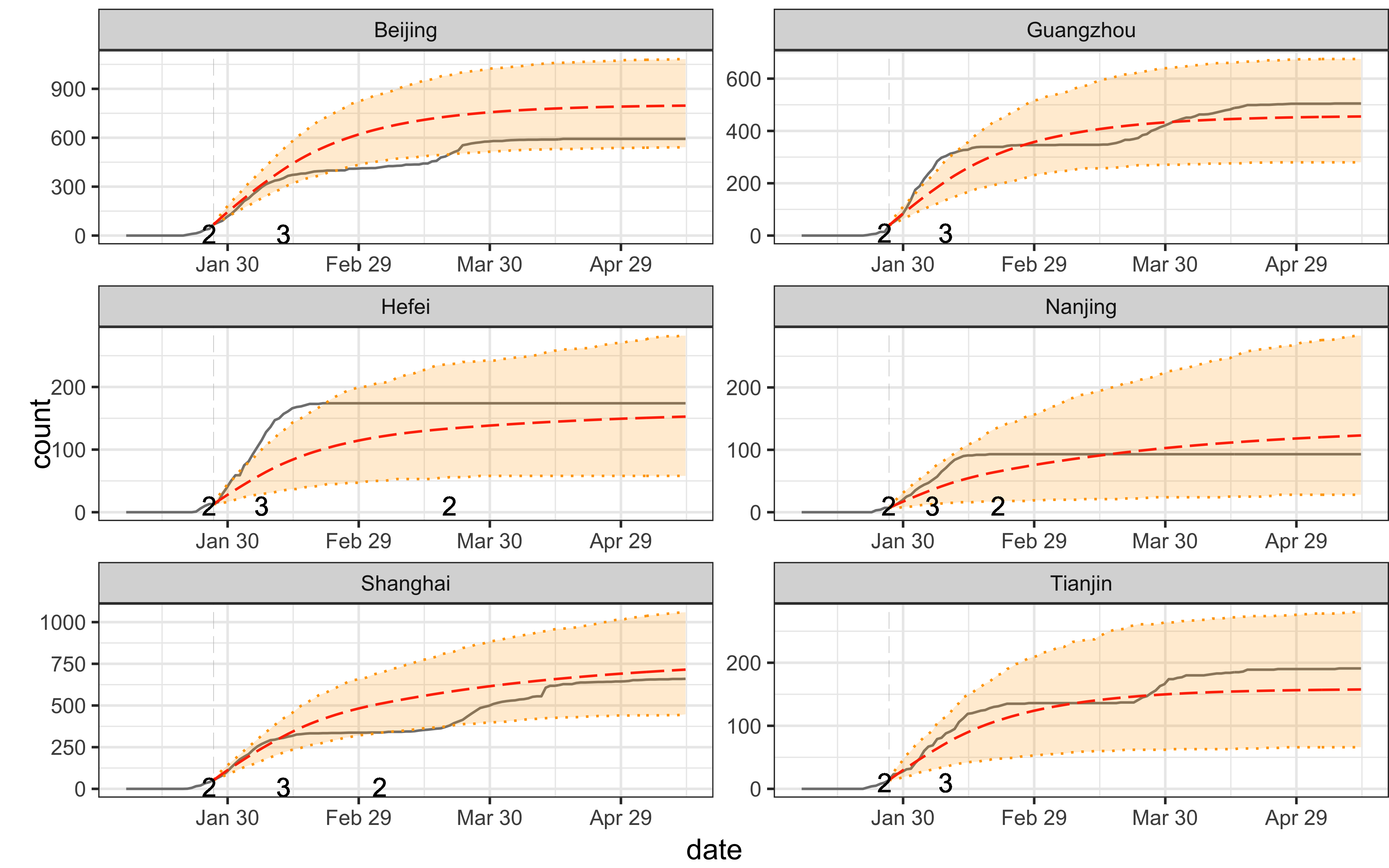} 
	\caption{Validation results for the six important cities in China. The solid lines are the observed counts of the cumulative infected cases. 
	The red dotted lines represent the mean predicted numbers, with the shaded areas indicating the $99\%$ prediction bands. 
	When a different action was taken, we annotate the new action level on the change point.}
	\label{figure:final_pred}
\end{figure}

The performance of our learned policies and the choices of weights both rely on the prediction accuracy of the estimated GSIR model, and we aim to first examine this point via temporal validation. 
Specifically, we first estimate the GSIR model using all data until day $12$,
and then for each city $l$, we predict $X^R_{l,t}$, the count of cumulative confirmed cases, from day $13$ to $120$ following the observed actions in data via forward sampling with the estimated GSIR model. 
Such a prediction can partially reflect the quality of our decisions suppose when we are on day $12$. 
The results aggregated over $1000$ replications are plotted in Figure \ref{figure:final_pred} and the prediction bands successfully cover the observed counts in most cases.

\begin{table*}[!t]
\centering
    \caption{Posterior means and standard deviations (in the parentheses) obtained using data for all the six cities until day $t_0$.}
    \label{table:1}
    \begin{tabular}{llllllll}
\toprule
  $t_0$ & $R_0^1$  &  $R_0^2$    & $R_0^3$  & 
  $\gamma$ &   $\beta_1$ &   $\beta_2$ &   $\beta_3$   \\
\midrule
12  & 4.13 (.34) & 0.74 (.09) & 0.37 (.06) & 0.07 (.005) & 0.29 (.011) & 0.05 (.005) & 0.03 (.004)  \\
61 & 2.32 (.09)  & 0.70 (.03) & 0.38 (.03) & 0.11 (.002) & 0.25 (.007) & 0.07 (.003) & 0.04 (.003) \\
\bottomrule
\end{tabular}
\end{table*}


To provide more insights into the intervention effects, in Table \ref{table:1}, we present the estimated parameters  using all data until day $61$, since the new cases afterwards are sparse. 
Here, $R_0^j = \beta_j / \gamma$ is the basic reproduction number under action level $j$, which plays an essential role in epidemiology analysis \cite{delamater2019complexity}. 
Roughly speaking, $R^j_0 < 1$ implies the pandemic will gradually diminish under action $j$. 
The small value of $R^3_0$ indicates that action $3$ has a significant effect on controlling the pandemic;
measure $2$ is also effective and is more like a mitigation strategy;
the value of $R^1_0$ emphasizes that a lack of intervention will lead to a disaster.  
The estimates of $R^1_0$ and intervention effects ($R^2_0 / R^1_0$ and $R^3_0 / R^1_0$) are consistent with the existing results surveyed in \cite{alimohamadi2020estimate} and \cite{cheatley2020effectiveness}. 
We also reported the estimates used to make the predictions in Figure \ref{figure:final_pred}. 
Although the estimation is not perfect due to data scarcity in the early stage, 
the prediction still captures the rough trend under a reasonable policy and we expect it will not affect the decision-making significantly.




\subsection{Evaluation of the Pareto-optimal policies}\label{sec:numerical_policy}
In this section, we conduct simulation experiments to compare the performance of our proposed method  with several competing policies on curbing the spread of COVID-19. 
Specifically, for the six cities, we start from day $12$ together, 
and follow the actions recommended by a specific policy until day $120$, with the cumulative costs recorded. 
The state transitions are generated from the GSIR model with  $\mathbb{E}(\rho_{61})$ as the parameter, and the observations before day $12$ are kept the same with the real data. 
This design is to mimic a real infectious disease control following some  specified policy or algorithm. 
Although this is a simulation study, we still investigated the validity of $\mathbb{E}(\rho_{61})$ as the environment parameter for fair comparison with the observed costs in the real data. 
Its validity is supported by the cross-validation results in Appendix \ref{sec:supp_val}. 
It also partially supports that these regions are similar. 

We compare our algorithm with several  expert policies that are motivated by real life observations and commonly considered in the literature \cite{merl2009statistical, lin2010optimal, ludkovski2010optimal}. 
Specifically, the comparisons are made among the following policies: 
\begin{enumerate}
    \setlength{\itemsep}{0pt}%
    \setlength{\parskip}{0pt}%
    \item Our proposed Pareto-optimal policies $\pi_k$ for $k \in \{-2, 0, \dots, 6\}$:  we fix the weight as $e^{k}/10$ across all cities and time, and follow the workflow proposed in Section \ref{sec:MDP}. $\mathcal{T}$ is set as every seven days. 
    We consider both the threshold-based policy class (\ref{eqn: policy_class_thre}) and the DQN-based method. 
    \item Occurrence-based mitigation policy $\pi^{M}_{m}$ for $m \in \{4, 6 \dots, 12\}$: a city implements action $2$ when there are new cases confirmed in the past $m$ days, and action $1$ otherwise.
    \item Occurrence-based suppression policy $\pi^{S}_{m}$ for $m \in \{4, 6, \dots, 10\}$: a city begins to implement action $2$ after $m$ days from its first confirmed case and strengthens it to level $3$ after $m$ more days. Afterwards, the city weakens the action to level $2$ when there have been no new cases for $m$ days and level $1$ there have been no new cases for $2m$ days.
    \item Count threshold-based policy $\pi^{TB}_{m}$ for $m \in \{5, 10, \dots, 50\}$: a city implements action $3$ when the count of new cases on the day before exceeds $m$, action $1$ when it is zero, and action $2$ otherwise.
    \item Behaviour policy $\pi^B$: the observed trajectories in the dataset.
\end{enumerate}

\begin{figure}[h!]
\centering		
\includegraphics[width=.55\textwidth, height=.6\textwidth]{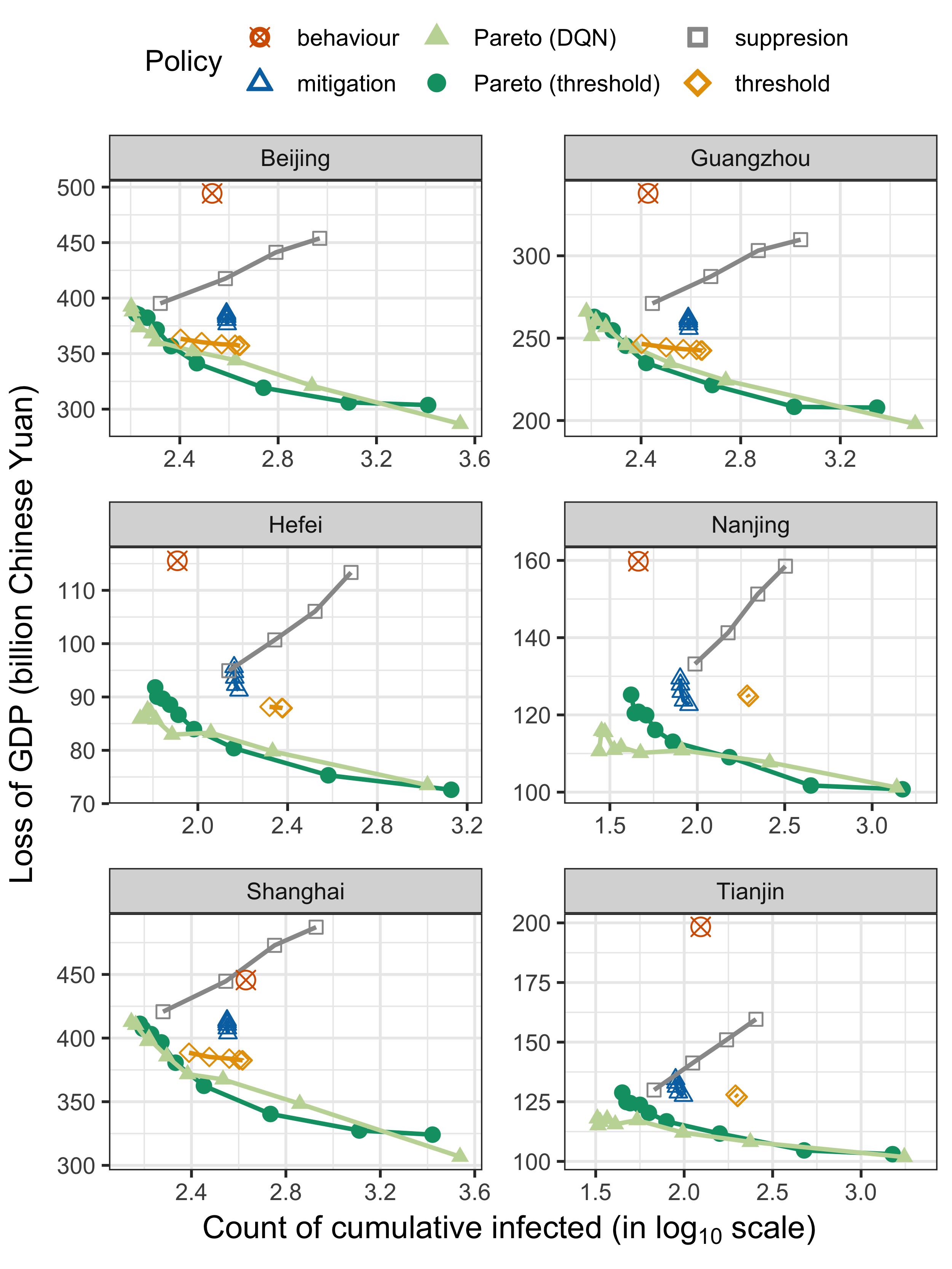} 
\caption{Cumulative epidemiological costs and economic costs  following different policies, averaged over $100$ replications. The closer to the left bottom corner, the better. The standard errors are negligible. 
On each curve, the different markers represent the performance of a policy with different hyper-parameters. 
}
\label{figure:Pareto}
\end{figure}

For each policy except for $\pi^B$, we run $100$ replications and present the average costs in Figure \ref{figure:Pareto}. 
We can see that the proposed method provides a clear view of the tradeoff and its performance is generally better than the competing  policies. 
In some cases, the DQN-based policies are slightly better than the threshold-based policies, while the overall value difference is not large. 
The behaviour policy by Chinese local governments is quite strict in the later period, and we interpret part of this effort as the attempt to curb the cross-border spread.
The other policies are not Pareto-optimal and also are not adaptive to  different situations in different cities. 
The clear trend among the occurrence-based suppression policies emphasizes the importance of intervening as promptly as possible, which can reduce both  costs. 

\setlength{\textfloatsep}{10.0pt plus 2.0pt minus 4.0pt}

To further investigate the robustness of the proposed method, two sensitivity analyses are conducted. 
Specifically, we repeat the above experiment with either various combinations of hyper-parameters or with the data generation model modified as a generalized Susceptible-Exposed-Infected-Removal model  \citep{tang2020review}. 
The results can be found in Appendix \ref{sec: supp-robust}. 
In reasonable ranges, the proposed method still yields consistent and superior performance. 

Finally, recall that infectious disease control is about trade-off. 
For a disease as infectious as the COVID-19, usually a stricter policy would be preferred. 
However, for a less infectious one, the same response policy might result in over-reaction and hence unnecessary costs to the economics. 
To demonstrate the adaptiveness of the proposed method, we repeat the above experiment with a less infectious disease. 
Specifically, we calibrate the GSIR model with the parameters of the 2009 H1N1 pandemic, with the other settings kept the same for ease of comparison. 
We choose the parameters estimated in \cite{fraser2009pandemic}, which reports an $R_0$ of about $1.4$. 

\begin{figure}[h!]
\centering		
\includegraphics[width=.55\textwidth, height=.6\textwidth]{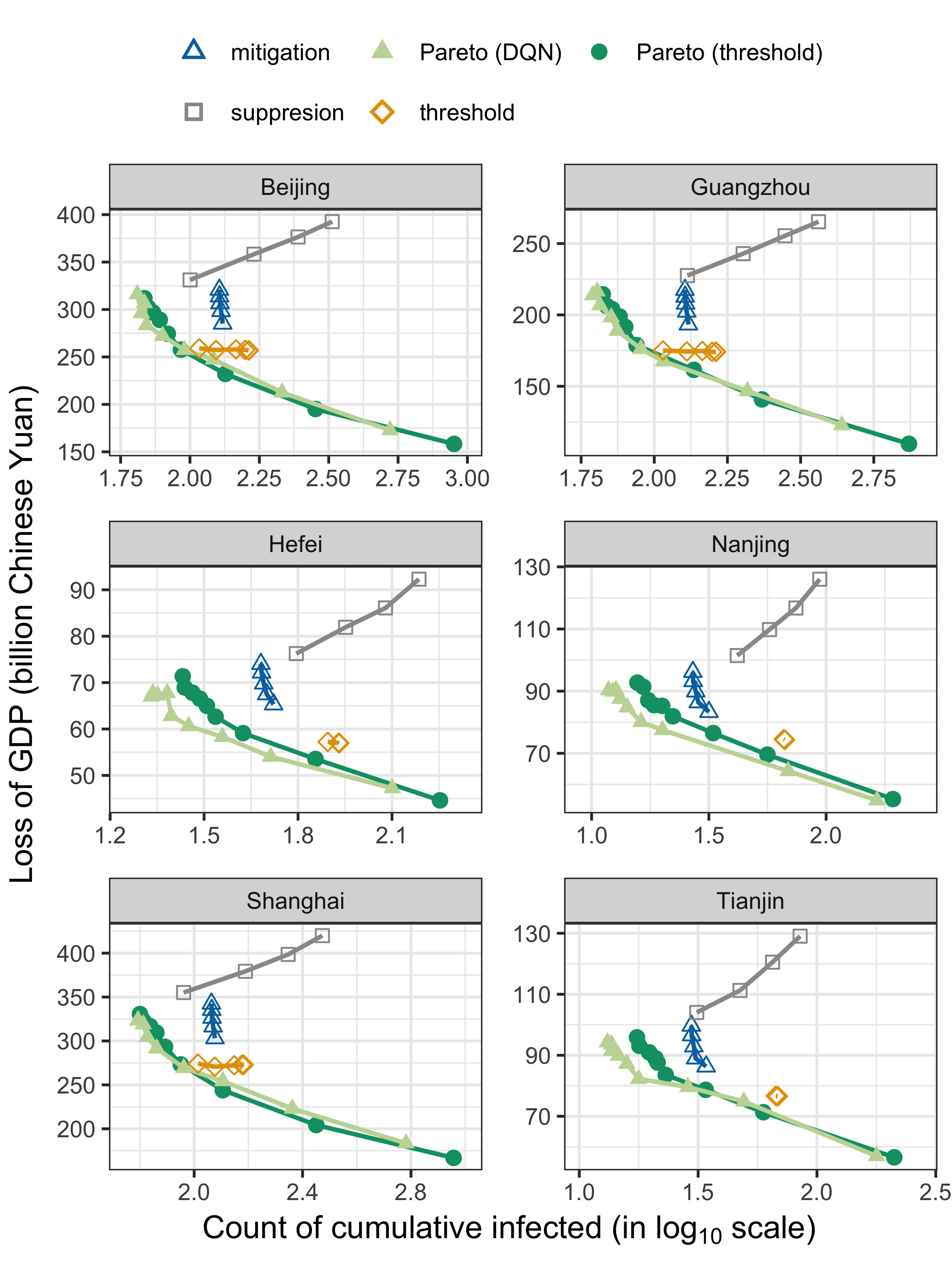} 
\caption{Experiment results with model parameters calibrated from the H1N1 pandemic. The costs of the behaviour policy are not displayed since they are not comparable.}
\label{figure:Pareto_H1N1}
\end{figure}

The results are summarized in Figure  \ref{figure:Pareto_H1N1}. 
We can see that the proposed method still yields superior performance. 
The count threshold-based policies achieve close performance with some Pareto-optimal policies, while our method provides more options with a clear trade-off. 
Compared to Figure \ref{figure:Pareto}, the economic costs of the strict  suppression policy increase, relative to the other policies. 

\section{Discussion}\label{sec:discussion}
Motivated by the ongoing COVID-19 pandemic and the witnessed challenges in decision-making for policymakers, 
this paper proposes a novel model-based multi-objective RL framework to assist policymakers in real-time decision-making with the objective of minimizing the overall long-term cost. 
The method shows promising performance in numerical studies. 

The overall framework is generally applicable to infectious disease pandemics. 
There are several components that can be extended: 
(i) other epidemiology models than the SIR model can also be used as the transition model, with the estimation method and policy class modified correspondingly;
(ii) more than two objectives can be similarly formalized, and resource limits can also be considered by introducing a cost to penalize exceeding the limits; 
(iii) other parametric policy classes can be similarly formalized depending on the information required for decision-making. 

We note that modeling the economic costs caused by various intervention measures is, admittedly, a challenging but important task. 
This is out of the scope of this paper and we use a simple proxy in the experiment for illustration purposes. 
A systematic analysis of the observed costs during COVID-19 would be a meaningful next step. 
Besides, choosing among the Pareto-optimal policies is another challenging task, which is arguably beyond the scope of data-driven methods. 
Options can be presented to a committee or the citizens for discussion or voting. 

As future directions, 
the spreads among multiple regions can be incorporated under the multi-agent RL framework, 
the heterogeneity of different regions can be accounted in a similar way as in \cite{qian2020lift} if we have enough data, 
and a multidimensional action space with the combinations of different interventions can be considered. 


Finally, we would like to emphasize that taking the economy as more  important than human lives is not a motivation or an outcome of this framework. 
On one hand, economic losses can also cause health damage to many people, probably no less than the direct damage from the disease; 
on the other hand, the estimated Pareto-optimal policies aim to help policymakers reduce the cost on one objective without sacrificing the other, as illustrated in Section \ref{sec:numerical_policy}.







\bibliographystyle{ACM-Reference-Format}
\bibliography{0_MAIN}

\clearpage

\appendix 

\raggedbottom
\section{Pseudo-code of the Policy Search Algorithms}\label{sec:pseudo_code}
In this section, we provide the pseudo-code for the two policy search algorithms introduced in Section 
\ref{sec:policy}. 
The hyper-parameters used in the experiments can be found in Appendix \ref{sec:supp-implementation}. 
In Algorithm \ref{alg:grid}, we describe the grid search based policy search algorithm when $J = 3$. 
In Algorithm \ref{alg:DQN}, we modify the deep-Q learning algorithm to take into consideration the learned uncertainty on $\vthe$.  
More details about the standard DQN can be found in \cite{mnih2015human}.

 \begin{algorithm}[!h]
\KwIn{bounds of the search space $u_2, u_3, U_2,  U_3$; step sizes $\xi_2, \xi_3$; number of replications $M$; 
data $\{ O^I_{l,t}\}_{t = t_0 - D}^{t_0}$ and $ \{A_{l,t}\}_{t = t_0 - D}^{t_0 - 1}$; 
other parameters $D, \rho_{t_0}, w_{l,t_0}, t_0, T, \mathcal{T}, c_l(\cdot)$}



set $\lam_2 = u_2$,  $V^* = +\infty$, $S_{l, t_0} =  MBS(\{ X^R_{l,t}\}_{t = t_0 - D}^{t_0}$, 
              $\{A_{l,t}\}_{t = t_0 - D}^{t_0 - 1}, \rho_{t_0}, D)$, and $X^R_{l,t} = O^I_{l,t}$ for $t \in \{t_0 - D, \dots, t_0\}$

\While{$\lam_2 \le U_2 $}{
    set  $\lam_3 = max(\lam_2, u_3)$
    
    \While{$\lam_3 \le U_3 $}{
        set $\vlam = (\lam_2, \lam_3)^T$ and the overall value $V = 0$
        
        \For{$m=1,\dots,M$}{
              
            \For{$t'=t_0, \dots,T$}{
              generate $S^G_{l, t'} =  MBS(
              \{ X^R_{l,t}\}_{t = t' - D}^{t'},$
              $\{A_{l,t}\}_{t = t' - D}^{t' - 1}, \rho_{t_0}, D)$

              choose action $A_{l,t'} {=} \pi(S^G_{l, t'}; \vlam) \mathbb{I}(t' {\in} \mathcal{T})
              + A_{l,t'-1} \mathbb{I}(t' {\notin} \mathcal{T})$

                sample $\vthe \sim \rho_{t_0}$,  $C^A_{l,{t'}} \sim c_l(A_{l,t'})$ 
                , and $S_{l,t'+1} \sim  f(\cdot|S_{l,t'}, 
                A_{l,t'}; \vthe)$
                
            
               calculate  $V = V + (X^S_{l,t'} - X^S_{l,t'+1} + \omega_{l,t_0} C^A_{l,{t'}})$  
             }  
        
        }
        \textbf{if} $V < V^*$ \textbf{then} update $V^* = V$ and $\vlam^* = \vlam$
        

        set $\lam_3 = \lam_3 + \xi_3$
    }
    
    set $\lam_2 = \lam_2 + \xi_2$
}
 \KwOut{ optimal policy $\hat{\pi}_{l, t_0}(\cdot; \omega_{l, t_0}) = \pi(\cdot; \vlam^*)$}
 \caption{Policy Search with Grid Search }\label{alg:grid} 
\end{algorithm}

\vspace{.2cm}
 \begin{algorithm}[!ht]
\KwData{
weight $\omega$, posterior $\rho_{t_0}$, current state $s_0$, current decision point $t_0$, horizon $T$
number of episodes $M$, exploration probability $\epsilon$
}

\textbf{Initialization:}
set the replay buffer $\mathcal{D}$ as empty, the action-value function $Q(\cdot, \cdot)$ with random weights

\For{episode $e = 1, \dots, M$}{
    randomly initialize with an appropriate state $s$
    
    \For{$t = t_0, \dots, T$}{
        with probability $\epsilon / ((e + 1) // 10 + 1)$ select a random action a; otherwise select $a = \argmin_a Q((s, t), a)$
        
        take action $a$, sample $\theta \sim \rho_{t_0}$, and then sample the next state $s'$ according to the GSIR model with parameter $\theta$
        
        sample the action cost for $a$ and calculate the weighted cost for weight $\omega$ as $c$
        
        store transition $(s, a, c, s', t)$ in $\mathcal{D}$
        
        sample a mini-batch $\{(s_i, a_i, c_i, s'_i, t_i)\}$ from $\mathcal{D}$
        
        set $y_i = c_i + min_a Q((s'_i, t_i + 1), a)$ if $s'_i$ is not terminal, and set $y_i = c_i$ otherwise
        
        perform a gradient descent step on  $\{((s_i, t_i), y_i)\}$
        
        set $s' = s$
    }
}

 \KwResult{
 optimal policy $\hat{\pi}_{t_0}(\cdot; \omega) = argmin_a Q((\cdot,t_0), a; \omega)$ and the recommended action 
 $\hat{\pi}_{t_0}(s_0; \omega)$
 }  
 \caption{Modified deep-Q learning with experience replay and epsilon-greedy exploration for finite-horizon planning.}\label{alg:DQN}
\end{algorithm}
\vspace{.5cm}


\vspace{-1cm}

\section{Robustness analysis results}\label{sec: supp-robust}
In Figure \ref{fig:robust}, we report the robustness analysis results. 
The experiment details can be found in Appendix \ref{sec:setting_robust}. 
In subplot (a), we present results when the true data generation process is a generalized Susceptible-Exposed-Infected-Removal model (SEIR) model. 
As expected, compared with results under the GSIR model, the performance of the  proposed method under this setting slightly deteriorates. 
However, in general, the method still over-performs the other fixed policies, and also provides more options with a clear trade-off.  

In subplot (b) and (c), we present results when different hyper-parameters are used. 
We present results for the threshold-based policy class. 
Results for the DQN-based methods are similar and omitted, for ease of presentation. 
These prior parameters turn out to have little importance in the long run, because their effects get washed out quickly by the observed data. 
The findings for the other prior parameters are similar. 


\begin{figure}[h]
\centering
\begin{subfigure}{0.75\textwidth}
\centering
\includegraphics[width=.8\textwidth, height=.58\textwidth]{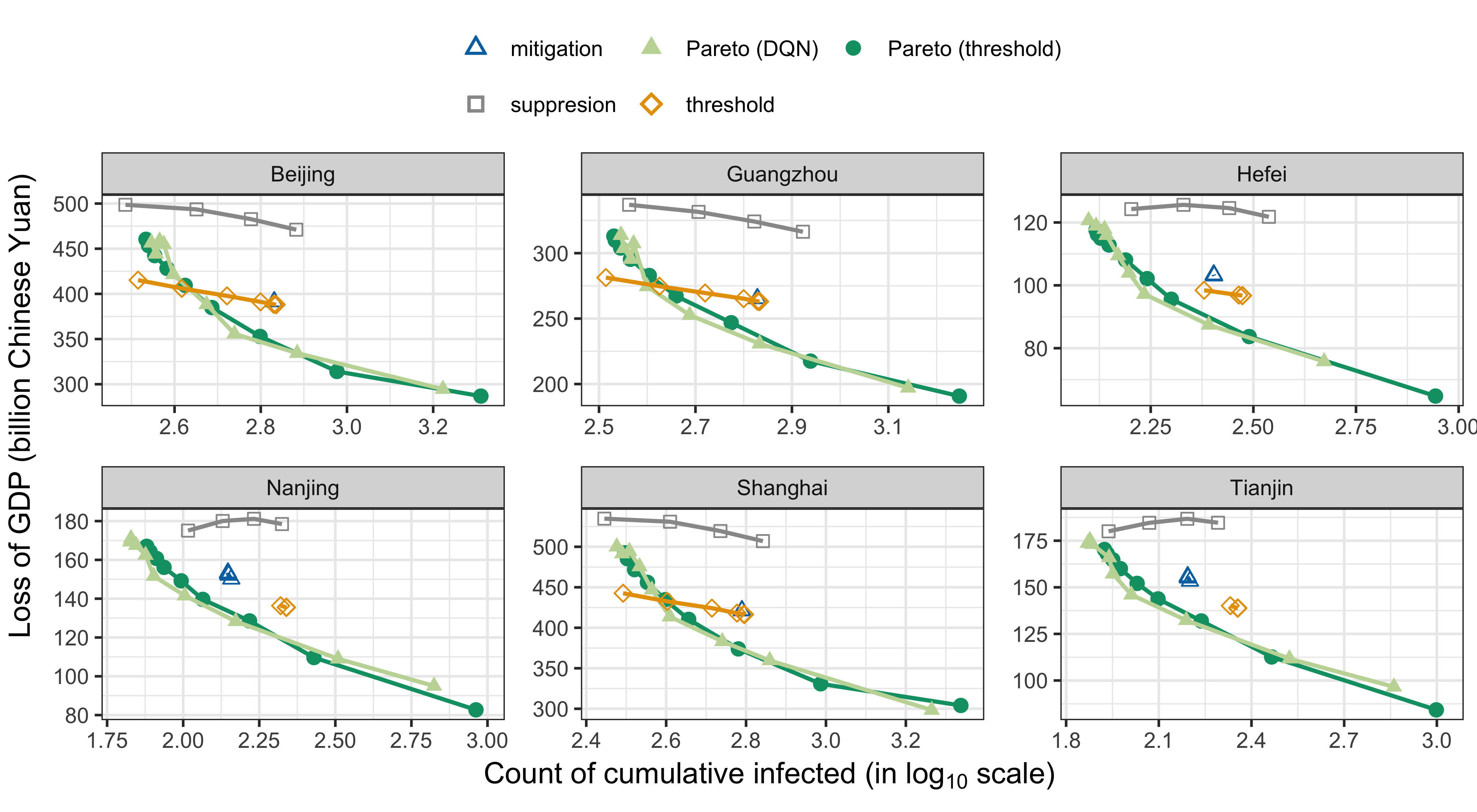} 
\caption{Results when the data generation model is the generalized SEIR model.}
\label{figure:Pareto_SEIR}
\end{subfigure}

\begin{subfigure}{0.75\textwidth}
\centering
\includegraphics[width=.8\textwidth, height=.58\textwidth]{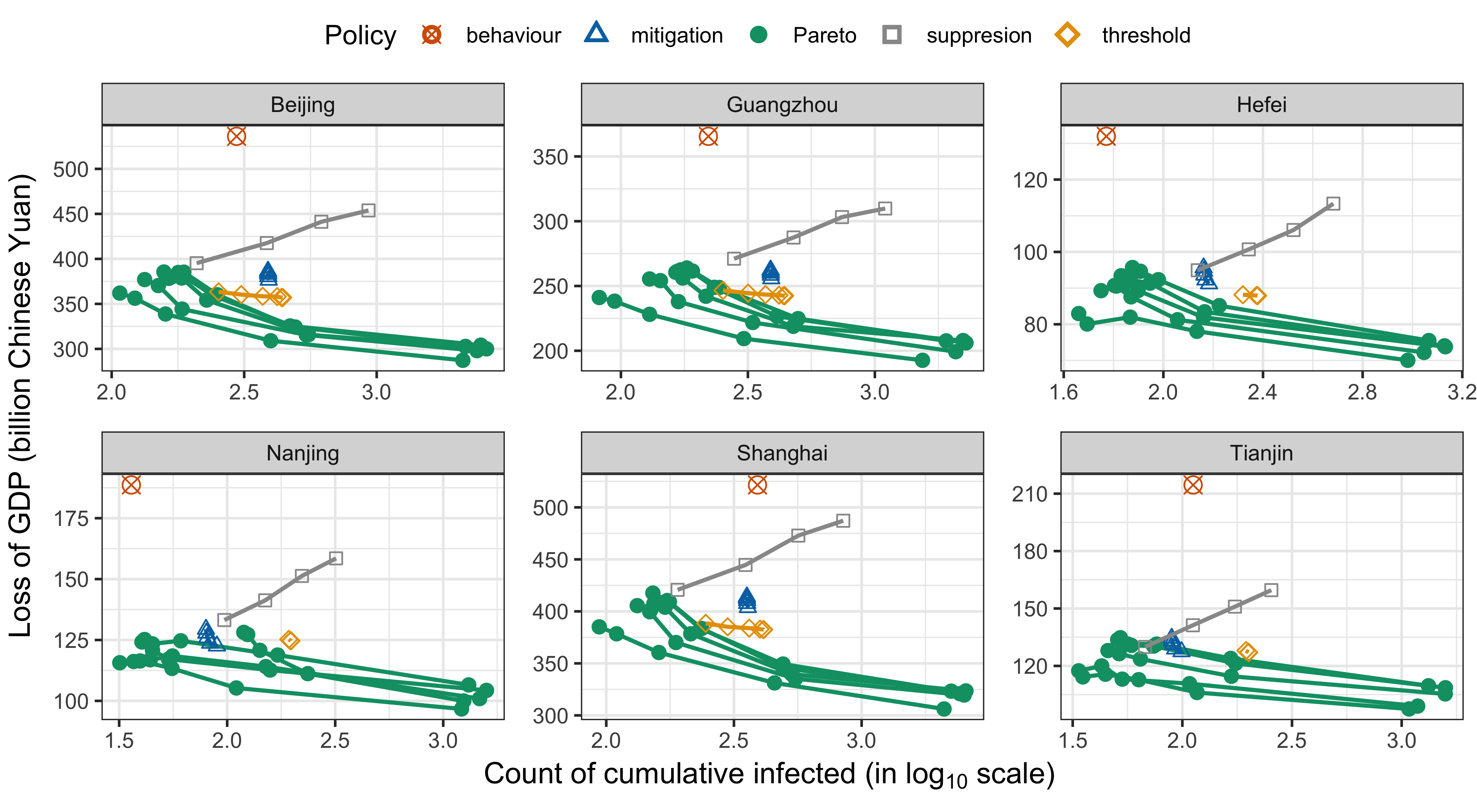} 
\caption{Results for different values of $D$.}
\label{figure:robust_lag}
\end{subfigure}

\begin{subfigure}{0.75\textwidth}
\centering
\includegraphics[width=.8\textwidth, height=.58\textwidth]{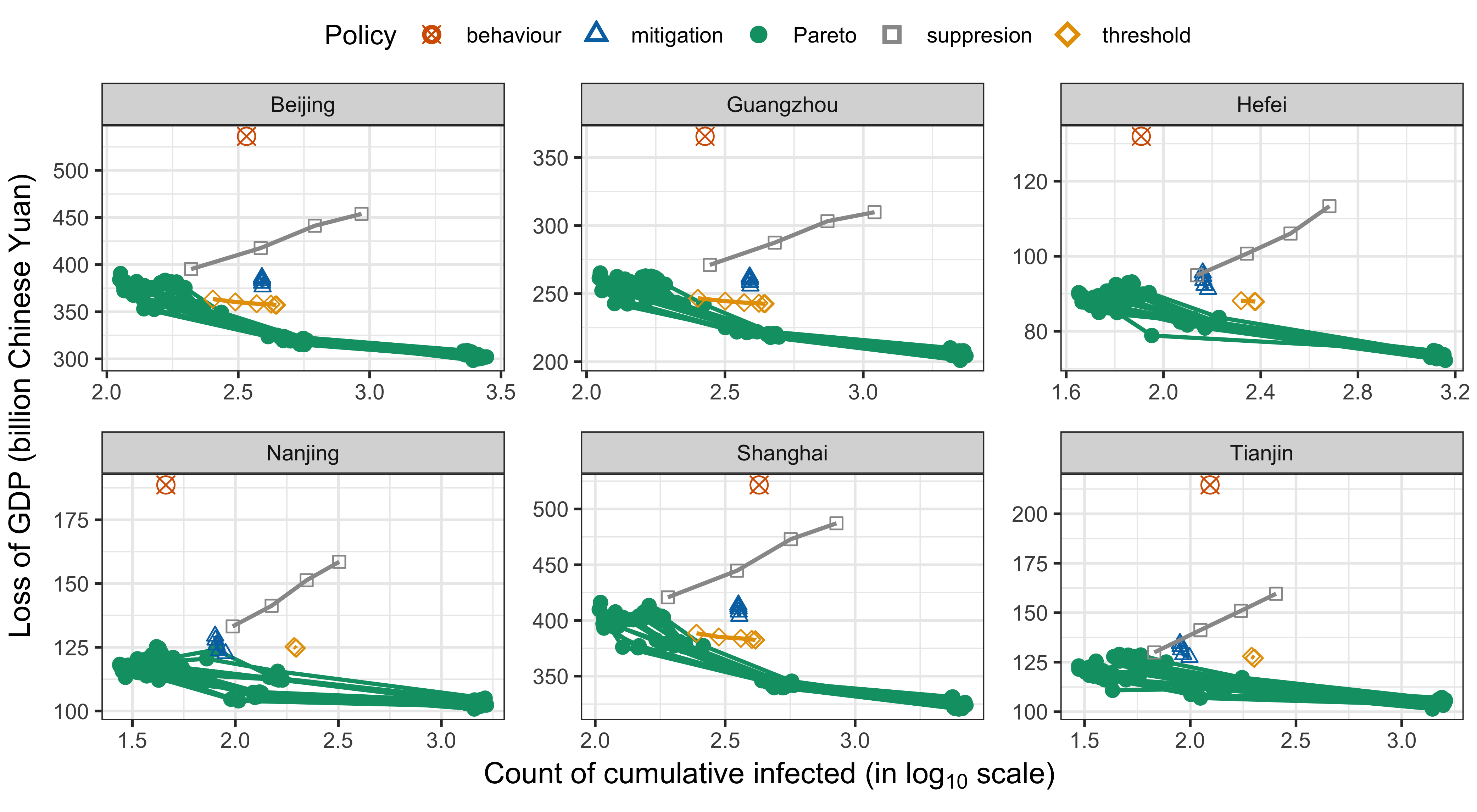} 
\caption{Results for different values of $\mu_1$ and $\mu_2$.}
\label{figure:robust_effect}
\end{subfigure}
\caption{Robustness analysis results}
\label{fig:sensitivity}
\end{figure}\label{fig:robust}








\section{Details of the experiments}


\subsection{Dataset and hyper-parameters}\label{sec:supp-data}
We obtain the counts of cumulative confirmed cases $O_{l,t}^I$ from  \cite{China_Data} with irregular records imputed by the average of their neighbors. 
For each region $l$, 
we collect its annual gross domestic product (GDP) $G_l$ and population $M_l$ from a Chinese demographic dataset \footnote{\url{https://www.hongheiku.com}}. 
The original mobility data collected from the online platform Baidu Qianxi\footnote{http://qianxi.baidu.com} includes ${m_{l,t}^{20}}$, the mobility of people in city $l$ on day $t$ in  2020, and ${m_{l,t}^{19}}$, the value for the same date in 2019, aligned by the lunar calendar.  
We define $r_{l,t}=1- m_{l,t}^{20}/m_{l,t}^{19}$ to 
measure the  ratio of human mobility loss due to $A_{l,t}$. 
The fitted distributions for the intervention cost function are $C_2 \sim \mathcal{N}(0.368, 0.239^2)$ and 
$C_3 \sim \mathcal{N}(0.484, 0.181^2)$.
As for the action data, we collect the date ranges for different control measures claimed by Chinese local governments. 
The date ranges for action $1$ and $2$ as well as the starting dates for action $3$ are collected from \cite{wiki:action}. 
The ending dates for action $3$ are manually collected from the local government websites or their social media accounts. 
We use the estimates for SARS \cite{mkhatshwa2010modeling} to set the priors for the disease features, specifically, we set $\gamma \sim Beta(178.89, 2000)$ and $ \beta_1 \sim Gamma(517.41, 2000)$.
As for $\beta_2$ and $\beta_3$, similar with the assumptions in \cite{ferguson2020impact}, we assume the two measures reduce the infection rate by $80\%$ and $90\%$ respectively, and therefore use $\beta_2 \sim Gamma(103.48 , 2000)$ and $ \beta_3 \sim Gamma(51.74, 2000)$.


\subsection{Implementation details}\label{sec:supp-implementation}

In our experiments, for the grid search algorithm, the number of rollout replications is set as $M = 100$. 
We run the grid search algorithm from coarse grids to finer grids to save computational cost. 
The schedule is listed in Table \ref{table:para}. 
In general, the larger the search spaces and the smaller the step sizes, the better the performance will be. 
In consideration of the importance of this application and the relatively lightweight computation, in real applications, we can choose the search spaces as large enough and step size as small enough. 
For DQN, we use a dense network of 5 layers, with 32 nodes per layer, to model the Q-function. 
The activation function is picked as relu and the optimizer is set as Adam. 
The learning rate is fixed as 0.0002 and the parameter for epsilon-greedy is set as $\epsilon = 0.1$. 

\begin{table}[h!]
\centering
    \caption{Search space and step sizes for the grid search in each round. $\lam^i_1$ and $\lam^i_2$ are the output from the $i$-th round.}
    \label{table:para}
\setlength\tabcolsep{2pt} 
    \begin{tabular}{lllllll}
\toprule
  Round & $u_2$  &  $u_3$    & $U_2$  & 
  $U_3$ &   $\xi_2$ &   $\xi_3$  \\
\midrule
1 & 0& 300& 50& 0& 100& 200 \\
2 &  max(0, $\lam^1_1$ - 50)&  $\lam^1_1$ {+} 50& 5& $\lam^1_2$ {-} 200& $\lam^1_2$ {+} 200& 20 \\
3 &  max(0, $\lam^2_1$ - 5)&  $\lam^2_1$ {+} 5& 1& $\lam^2_2$ {-} 20& $\lam^2_2$ {+} 20& 4 \\
4 &  max(0, $\lam^3_1$ - 1)&  $\lam^3_1$ {+} 1& 0.25& $\lam^3_2$ {-} 4& $\lam^3_2$ {+} 4& 1 \\
5 &  max(0, $\lam^4_1$ - 0.25)&  $\lam^4_1$ {+} 0.25& 0.01& $\lam^4_2$ {-} 1& $\lam^4_2$ {+} 1& 0.2 \\
\bottomrule
\end{tabular}
\end{table}

\vspace{-.2cm}

\subsection{Validity of the simulation environment}\label{sec:supp_val}

We apply a leave-one-out cross-validation procedure to show that our simulation environment is close to the real data generation process. 
Specifically, for $l = 1, \dots, 6$, we follow the steps: 
\begin{enumerate}
    \item estimate the environmental parameter $\vthe$ as $\hat{\vthe}_l$, using all data  until day $61$ of cities in $\{1, \dots, 6\} \setminus \{l\}$.
    \item for city $l$, based on $\hat{\vthe}_l$, we predict its counts of the cumulative infected cases from day $12$ to day $120$, following the observed actions in the data. Denote the estimate for day $120$ as $\hat{X}^R_{120, l}.$ 
    \item calculate the approximation error by $\epsilon_{l} = |\hat{X}^R_{120, l} - O^I_{120,l}| / O^I_{120,l}$, where $O^I_{120,l}$ is the observed cumulative infected count for city $l$ until day $120$. 
\end{enumerate}

The average error ratio $\sum_{l = 1}^6 \epsilon_{l} / 6 = 0.18$  indicates that the environment is a relatively valid proxy for the real data generation process, which supports the fairness of our comparison with the behaviour policy. 
It also partially supports that these regions are similar.

\subsection{Setting for the robustness analysis}\label{sec:setting_robust}
The Susceptible-Exposed-Infected-Removal model (SEIR)  model is another commonly used compartmental model, which adds one additional exposed compartment to the SIR model \citep{tang2020review}.  
We generalize the SEIR model in a similar with as the GSIR model, by including the action effects into consideration to set the infection rate $\beta$. The transition from the susceptible compartment to the exposed compartment is assumed to follow a binomial distribution with parameter $\alpha$. 
According to \cite{yang2020modified}, we set the incubation rate $\alpha$ as $1/7$. 
The other parameters are kept the same with Section \ref{sec:numerical_policy}. 
Since the exposed compartment is typically latent,  the number of individuals in this compartment on the beginning date can not be set based on the real data as the other compartments. 
For experiment purpose, we set the number to be the same as the number in the infectious compartment. 

For sensitiveness to hyper-parameters, we considered two settings: 
(1) $D = 7, 8, 9, 10, 11$, with the other hyper-parameters fixed; 
(2) $\{(\mu_2, \mu_3): \mu_3 \in \{0.05, 0.1, 0.15, 0.2\}
, \mu_2 - \mu_3 \in \{0.05, 0.1, 0.15, 0.2\} )\}$, with the other hyper-parameters fixed. Recall that we define $\mu_j = \beta_j / \beta_1$ and use it to set the prior for $\beta_i$. 
The other settings are kept the same with Section \ref{sec:numerical_policy}, except for that we only run 25 replications for the proposed algorithm to save computation. 



\end{document}